%% file: main.tex
\documentclass{article} 
\usepackage{iclr2027_conference,times}

\usepackage[utf8]{inputenc} 
\usepackage[T1]{fontenc}

\usepackage{mathtools}      
\usepackage{amssymb}        
\usepackage{amsthm}
\usepackage{nicefrac}

\usepackage{graphicx}
\usepackage{wrapfig}
\usepackage{rotating}
\usepackage{subcaption}

\usepackage{array}
\usepackage{booktabs}
\usepackage{longtable}
\usepackage{multirow}
\usepackage{makecell}
\usepackage{diagbox}

\usepackage{arydshln}

\usepackage{microtype}
\usepackage{xcolor}
\usepackage{enumitem}

\usepackage{hyperref}
\usepackage{url}

\newcolumntype{C}[1]{>{\centering\arraybackslash}p{#1}}
\newcolumntype{L}[1]{>{\raggedright\arraybackslash}p{#1}}

\newcommand{\whitebox}{\ensuremath{\circ}}
\newcommand{\blackbox}{\ensuremath{\bullet}}

\newcommand{\eg}{\textit{e.g.}}

\title{
From Textual Outputs to Probabilistic Signatures: Robust Black-Box Model Ownership Verification

}

\author{%
\begin{minipage}[t]{\dimexpr\textwidth-2\tabcolsep\relax}
\normalfont\raggedright
\textbf{Xiaokun~Yang}\textsuperscript{1,2}\quad
\textbf{Jian~Liang}\textsuperscript{1,2,}\thanks{Corresponding author.}\quad
\textbf{Anda~Cheng}\textsuperscript{3}\quad
\textbf{Jun~Feng}\textsuperscript{1,2}\quad
\textbf{Yesheng~Liu}\textsuperscript{1,2}\\[3pt]
\textbf{Yinggui~Wang}\textsuperscript{3}\quad
\textbf{Tao~Wei}\textsuperscript{3}\quad
\textbf{Ran~He}\textsuperscript{1,2}\quad
\textbf{Tieniu~Tan}\textsuperscript{1,2,4}\\[8pt]
\textsuperscript{1}\,NLPR \& MAIS, Institute of Automation,
Chinese Academy of Sciences\\
\textsuperscript{2}\,School of Artificial Intelligence,
University of Chinese Academy of Sciences\\
\textsuperscript{3}\,Ant Group\\
\textsuperscript{4}\,Nanjing University\\[4pt]
\texttt{yyyxkun02@gmail.com, liangjian92@gmail.com}
\end{minipage}%
}

\iclrfinalcopy 
\begin{document}

\maketitle
\lhead{Under review}

\begin{abstract}

  The substantial computational resources required to train large language models (LLMs) make them valuable intellectual property, raising concerns about model theft and unauthorized deployment.
  Fingerprinting provides a practical approach to ownership verification by determining whether a suspicious model is derived from a protected source model.
  However, existing black-box methods often fail to provide reliable verification, as textual signals are unstable under stochastic sampling.
  This issue is pronounced for reasoning models, where long-form generation amplifies the variability introduced by sampling.
  Although sampled texts vary, the underlying probability patterns remain relatively stable.
  Motivated by this insight, this work proposes Token-Level Probability Fingerprinting (\texttt{TOPF}), a robust framework that leverages probability distributions as more stable verification signals than textual outputs.
  Since derived models tend to preserve probability patterns similar to their sources, \texttt{TOPF} extracts discriminative features by evaluating the token-level probabilities assigned by the source model to responses generated by a suspicious model.
  Extensive experiments on both reasoning models and standard LLMs demonstrate that \texttt{TOPF} achieves effective verification performance approaching white-box methods.
  Code is available at \url{https://github.com/YyyxKun/TOPF}.
\end{abstract}

\section{Introduction}


Large language models (LLMs) excel at complex problem solving, but their development demands substantial computation and high-quality data, making them valuable intellectual property~\citep{li2023protecting}.
This value has raised concerns over copyright infringement and model theft~\citep{carlini2024stealing,oliynyk2023survey}. 
Adversaries may exploit proprietary models through illegal deployment or unauthorized API wrapping, harming developers and the broader AI ecosystem~\citep{shao2025sok}.

\begin{wrapfigure}[15]{r}{0.60\textwidth}
\centering
\includegraphics[width=\linewidth]{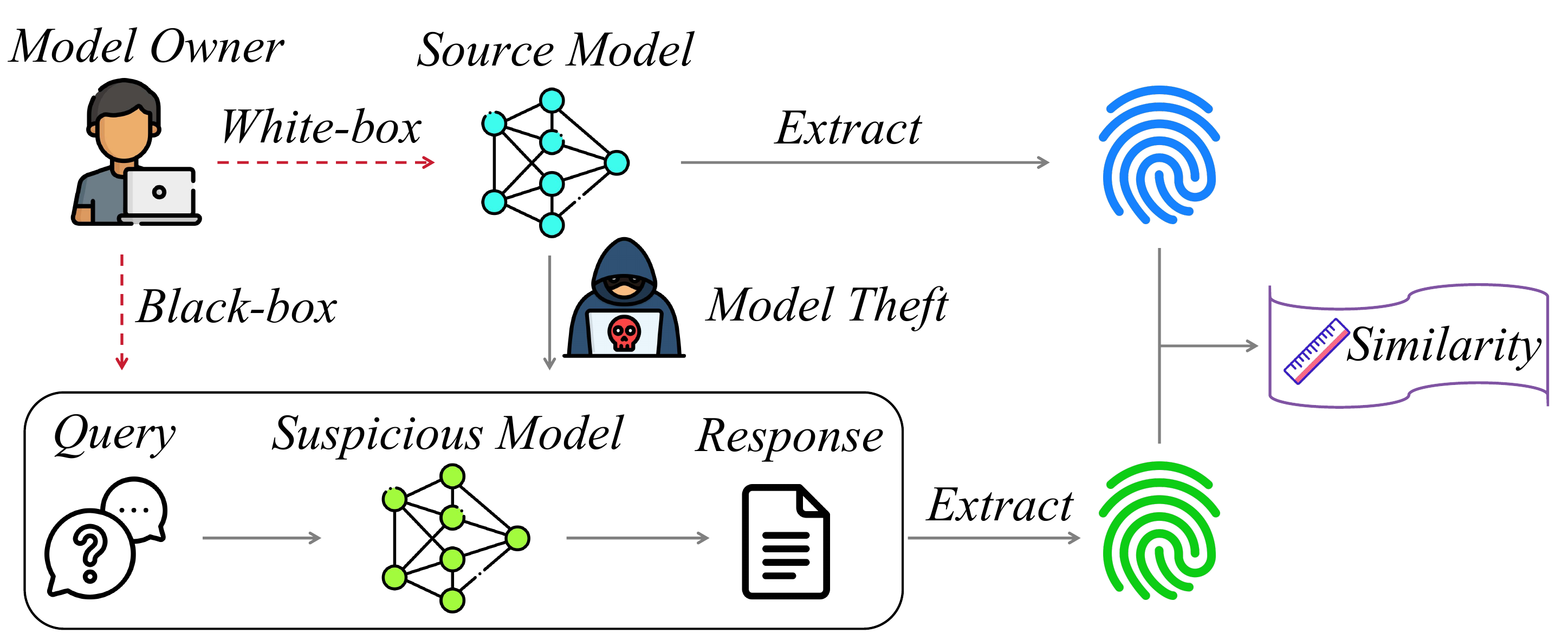} 
\caption{Illustration of intrinsic fingerprinting for ownership verification. Fingerprints are extracted from the protected source model with white-box access, while the suspicious model is examined only through black-box queries.
}
\label{fig:illustration}
\end{wrapfigure}

To address these concerns, fingerprinting has been proposed for ownership verification~\citep{guan2022you} as shown in Figure~\ref{fig:illustration}. 
Existing approaches fall into two categories: invasive and intrinsic~\citep{chen2022copy}.
Invasive fingerprinting, also known as model watermarking, embeds predefined triggers or patterns into models during training or fine-tuning~\citep{liu2026distilling,yamabe2025mergeprint}, which alters model parameters and may harm utility and generalization.
In contrast, intrinsic fingerprinting identifies models through their inherent behaviors~\citep{xu2025rap}.
Depending on the level of access to the suspicious model, intrinsic methods are further divided into white-box and black-box settings~\citep{shao2025reading}. 
In this work, we focus on black-box intrinsic fingerprinting, which represents a practical yet challenging setting for verification.

However, existing black-box methods often provide unreliable verification because they rely on textual signals that are unstable under stochastic sampling.
These methods typically compare generated outputs or optimize prompts to elicit predefined responses~\citep{jin2024proflingo,iourovitski2024hide}, both of which require stable output behavior~\citep{zhu2025auditing,yan2025duffin}.
Stochastic sampling weakens this stability by allowing the same query to yield different responses.
This issue is particularly pronounced in reasoning models~\citep{wang2025comprehensive,xu2025large}, whose long-form reasoning trajectories amplify variability induced by sampling across generation steps.

To address these challenges, we propose Token-Level Probability Fingerprinting (\texttt{TOPF}), a robust method that leverages probabilistic behavior.
Our approach is grounded in two insights: (1) probability distributions provide more stable signals than textual outputs~\citep{mitchell2023detectgpt}, and (2) derived models tend to preserve probability patterns similar to their sources~\citep{tong2025seedprints}.
Specifically, \texttt{TOPF} reframes ownership verification to a relative comparison task using public reference models as anchors.
In our method, we query the suspicious model and compute token-level probabilities of its responses under both the source and reference models to obtain probability matrices. 
These matrices are then encoded into compact representations. 
Finally, ownership is determined by whether the suspicious model is relatively closer to the source model than to the reference models.

Extensive experiments show that \texttt{TOPF} achieves effective verification performance on both reasoning models and standard LLMs. 
The framework also demonstrates strong robustness to different reference set sizes and query settings. 
These results indicate that \texttt{TOPF} provides a reliable solution for ownership verification in practical black-box scenarios. 

In summary, our main contributions are:
\begin{itemize}[left=10pt]
    \item We identify stochastic sampling as a key challenge for black-box methods and show that textual signals become unreliable as the sampling temperature increases.
    \item We propose \texttt{TOPF}, a relative comparison framework that leverages stable probabilistic signals underlying generated text to quantify model similarity.
    \item We conduct extensive experiments to validate the effectiveness of \texttt{TOPF} on both reasoning models and LLMs, demonstrating performance comparable to white-box methods.
\end{itemize}

\section{Related Work}

\paragraph{Invasive Fingerprinting.}
Invasive fingerprinting, also known as model watermarking~\citep{shao2025explanation}, embeds ownership information into model parameters, typically by inducing backdoor-like behaviors through fine-tuning or distillation.
\textbf{Pair-based fingerprinting} encodes ownership by training the model to memorize query-response pairs.
IF~\citep{xu2024instruct} and SF~\citep{nasery2025scalable} follow this paradigm via regularized fine-tuning, enabling verification while preserving model utility.
FPEdit~\citep{wang2025fpedit} further localizes information injection through model editing.
In contrast, \textbf{distribution-based fingerprinting} avoids fixed query-response pairs and instead embeds ownership signals within the output distribution~\citep{gloaguen2025llmfingerprinting}.

\paragraph{Intrinsic Fingerprinting.}
Intrinsic fingerprinting assumes that ownership information is inherently encoded in a trained model without explicit intervention.
\textbf{White-box fingerprinting} requires access to internal information such as model weights or representations.
HuRef~\citep{zeng2024huref} builds invariant terms from the stable directions of parameter vectors, and REEF~\citep{zhang2025reef} compares internal representations with centered kernel alignment.
\textbf{Black-box fingerprinting} verifies model identity solely by querying the target model and analyzing its outputs.
LLMmap~\citep{pasquini2025llmmap} uses carefully designed prompts to elicit distinctive response patterns.
MET~\citep{gao2025model} formulates fingerprinting as a statistical test for detecting distributional discrepancies between models, whereas TRAP~\citep{gubri2024trap} employs GCG-based adversarial suffixes to trigger predefined responses, enabling identity verification with few queries.

\section{Threat Model}

\paragraph{Problem Formulation.}
Model ownership verification aims to assess whether a suspicious model $M_t$ is derived from a source model $M_s$~\citep{xiong2025iseal, li2023plmmark}. 
A model $M_t$ is considered derived from $M_s$ if it is obtained through a transformation $\mathcal{T}$, such as fine-tuning, quantization, or model merging, i.e., $M_t=\mathcal{T}(M_s)$. 
By contrast, an independent model $M_{\mathrm{ind}}$ is trained separately from $M_s$ and is not derived from it.
The goal is to design a verification function $V(M_s,M_t)$ that outputs a similarity score indicating whether $M_t$ is derived from $M_s$~\citep{li2022fedipr}. 

\paragraph{Adversary's Goal.}
The adversary aims to appropriate the capabilities of $M_s$ while concealing its provenance. 
Specifically, the adversary seeks to preserve the knowledge of $M_s$, modify the model to evade simple similarity checks, and claim independent authorship or ownership of $M_t$. 
We assume the adversary may apply arbitrary post-processing or deployment strategies, but has no access to the model owner's secret verification mechanisms.



    

\paragraph{Model Owner's Goal and Capability.}
The model owner aims to reliably determine whether a suspicious model $M_t$ is derived from the source model $M_s$, remaining effective under substantial modifications while avoiding false accusations against independently trained models.
To this end, we assume the model owner possesses the following capabilities:
\begin{itemize}[left=10pt]
    \item \textbf{White-box access to the source model.}  
    The model owner has full access to $M_s$, including its architecture and parameters, and can perform unrestricted analysis or computation on the model.

    \item \textbf{White-box access to reference models.}
    The model owner may collect open-source models as reference anchors to support verification decisions.
    This assumption is realistic since such models and their weights are widely available.
    
    \item \textbf{Black-box access to the suspicious model.}  
    The model owner can only interact with $M_t$ through query-response APIs, without access to its internal parameters or architecture.  
\end{itemize}

\section{Methodology}

\begin{wrapfigure}[16]{r}{0.6\textwidth}
\centering
\includegraphics[width=\linewidth]{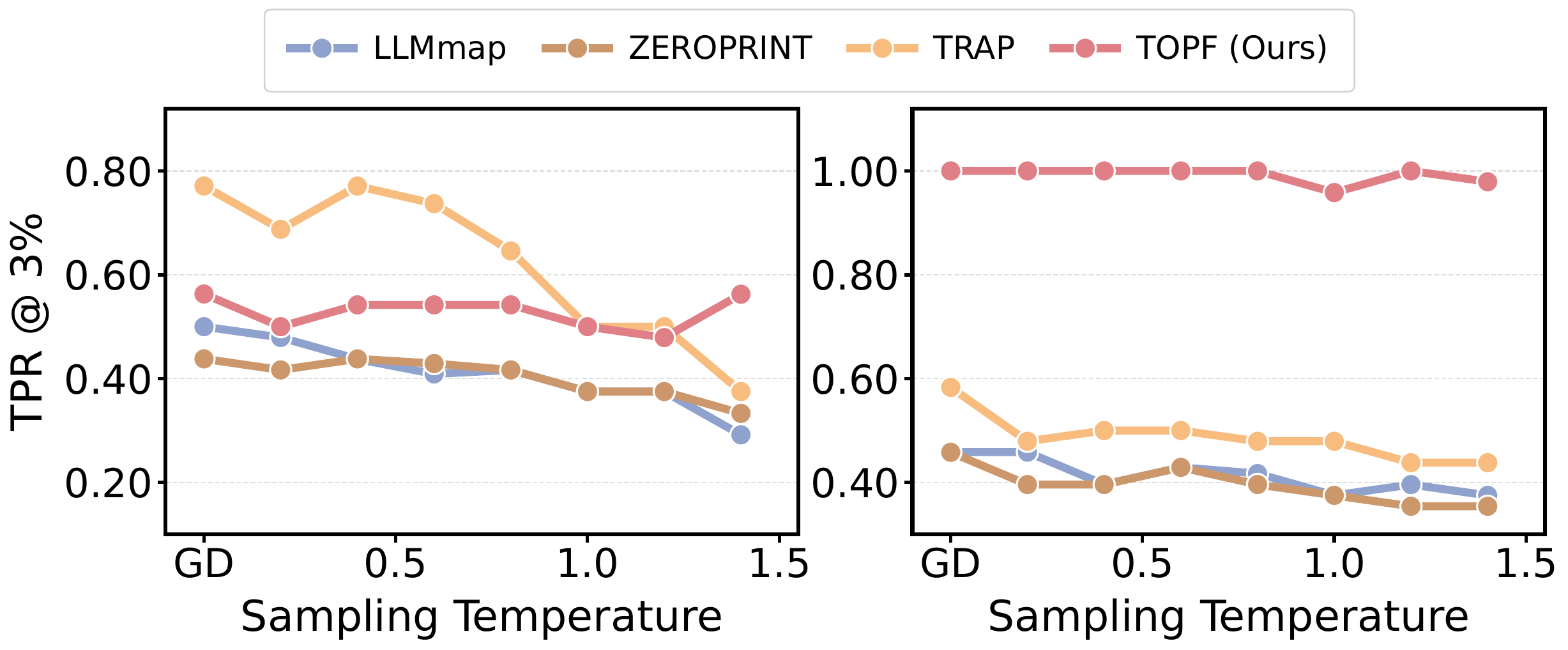} 
\caption{Performance across different sampling temperatures on Llama3.1-8B and Qwen2.5-7B. GD denotes greedy decoding. As temperature increases, baseline methods degrade, whereas \texttt{TOPF} remains consistently robust.}
\label{fig:temp}
\end{wrapfigure}

Existing black-box methods are vulnerable to stochastic sampling, since they rely on textual signals that may change across different decoding outcomes.
As shown in Figure~\ref{fig:temp}, increasing the sampling temperature degrades prior methods, indicating that direct output comparison becomes unreliable when generation paths become diverse.
Motivated by the insight that probability patterns remain more stable than textual outputs, we propose Token-Level Probability Fingerprinting (\texttt{TOPF}) for black-box ownership verification. 
Unlike prior approaches that directly measure similarity between the source model $M_s$ and the suspicious model $M_t$, \texttt{TOPF} introduces a set of reference models $\mathcal{F}=\{M_f^1,\ldots,M_f^R\}$ as anchors, enabling verification from a comparative perspective.

\begin{figure}[t]
\centering
\includegraphics[width=\linewidth]{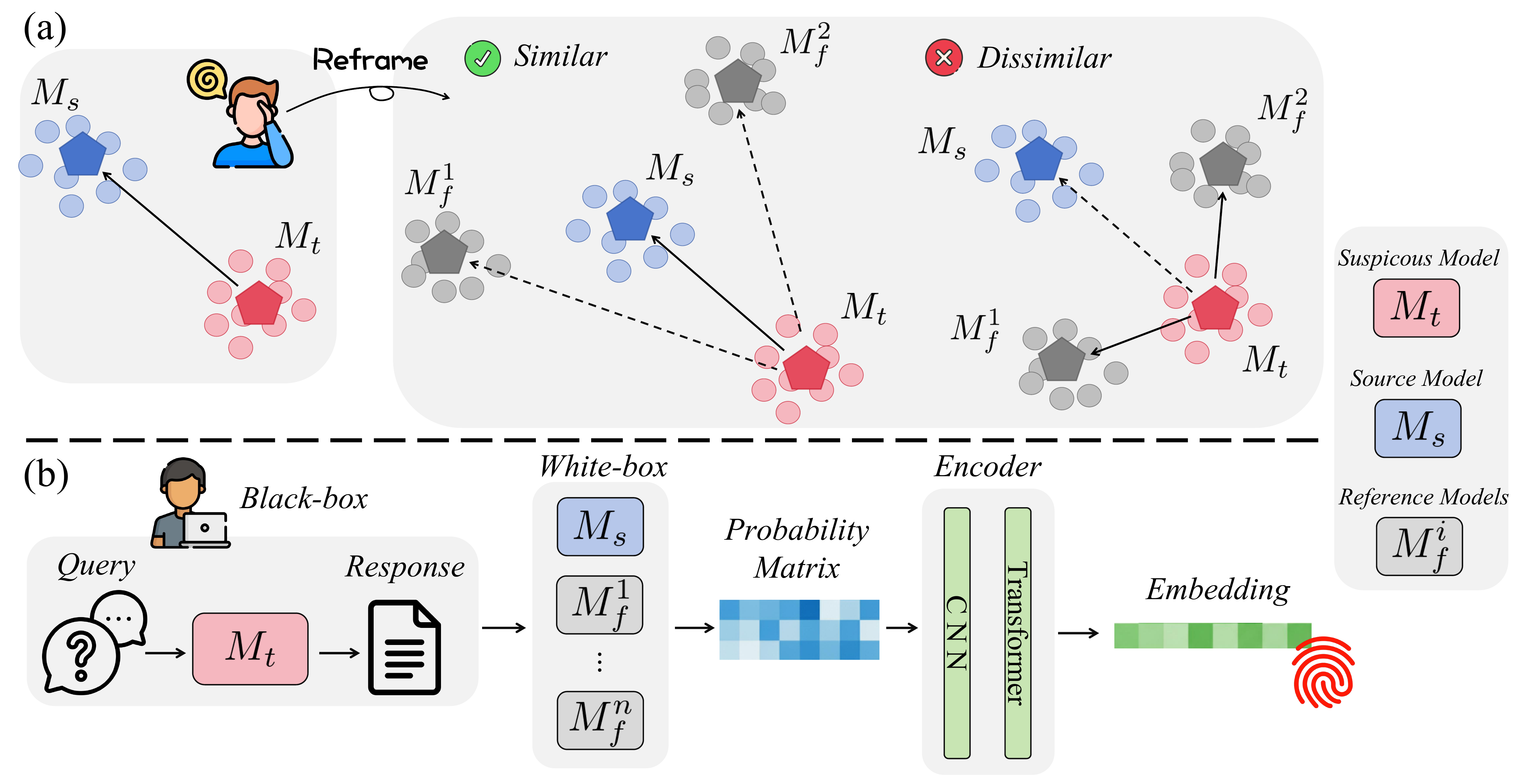}
\caption{
\textbf{Overview of Token-Level Probability Fingerprinting (\texttt{TOPF}).}
(a) \texttt{TOPF} reframes ownership verification as relative comparison by introducing open-source reference models $\{M_f^i\}$. A suspicious model is considered derived only if it is relatively closer to the source model $M_s$ than to the reference models.
(b) Fingerprints extraction pipeline. Responses from the suspicious model are evaluated by $M_s$ and $\{M_f^i\}$ to obtain token-level probabilities, which are stacked into a matrix and encoded into a compact embedding for verification.
}
\label{fig:pipeline}
\end{figure}

\subsection{Token-Level Probability Trace Extraction}

We construct a prompt pool $\mathcal{X}=\{x_i\}_{i=1}^{N}$ consisting of open-source prompts that cover diverse domains, including logical reasoning, code generation, and mathematical problem solving.
For each model $M_m$ in the model set $\mathcal{M}=\{M_s\}\cup\mathcal{F}$, we obtain a response $y_i^{(m)}$ for each prompt.
We compute the token-level log probabilities of each response under all models in $\mathcal{M}$.
For a model $M_j \in \mathcal{M}$, the corresponding probability trace is defined as:
\begin{equation}
\mathbf{p}_{i}^{(m,j)} =
\left[
\log P_{M_j}\!\left(y_{i,t}^{(m)} \mid x_i, y_{i,<t}^{(m)}\right)
\right]_{t=1}^{T}.
\end{equation}

Each trace is aligned to a fixed length $T$ via padding or truncation.
It characterizes how $M_j$ evaluates the step-wise generation behavior of a response produced by $M_m$.
We stack the probability traces across all models in $\mathcal{M}$ to construct a probability matrix:
\begin{equation}
\mathbf{P}_{i}^{(m)} = \left[ \mathbf{p}_{i}^{(m,j)} \right]_{M_j \in \mathcal{M}} \in \mathbb{R}^{|\mathcal{M}| \times T}.
\end{equation}
This matrix captures how a response generated by $M_m$ is probabilistically evaluated by models from different sources, and serves as the fundamental representation for subsequent verification.

\subsection{Model-Specific Representation Learning}

\paragraph{Encoder Architecture.}
The encoder is the core module for fingerprints extraction, mapping the raw probability matrix $\mathbf{P}_{i}^{(m)} \in \mathbb{R}^{|\mathcal{M}|\times T}$ to a discriminative latent representation.
We employ a hybrid architecture that combines the local feature extraction capabilities of CNNs with the global context awareness of Transformers~\citep{wang2023seqxgpt}.

Specifically, stacked 1D convolutional layers first operate along the temporal dimension to reduce local noise and extract sequential patterns. The resulting feature maps are then fed into Transformer blocks, where self-attention models long-range dependencies across the reasoning chain. A pooling layer is applied to the final Transformer output to produce the fixed-dimensional embedding $\mathbf{z}_{i}^{(m)}$:
\begin{equation}
\mathbf{z}_{i}^{(m)} = f_{\theta}(\mathbf{P}_{i}^{(m)}),
\end{equation}
where $f_{\theta}(\cdot)$ denotes the CNN-Transformer encoder. Model architecture and training details are provided in Appendix~\ref{app:encoder} and Appendix~\ref{app:encoder_train}.

\paragraph{Training Objective.}
We adopt a supervised contrastive objective, where embeddings from the same model are treated as positives and those from different models as negatives.
For an anchor $\mathbf{z}_{i}^{(m)}$, let $\mathcal{P}_{i}^{(m)}$ denote its positive set in the mini-batch. We minimize:
\begin{equation}
\mathcal{L}_{i}^{(m)} =
-\frac{1}{|\mathcal{P}_{i}^{(m)}|}
\sum_{\mathbf{z}^{+}\in\mathcal{P}_{i}^{(m)}}
\log
\frac{
\exp\!\left(\mathrm{sim}(\mathbf{z}_{i}^{(m)},\mathbf{z}^{+})/\tau\right)
}{
\sum\limits_{\mathbf{z}\neq \mathbf{z}_{i}^{(m)}}
\exp\!\left(\mathrm{sim}(\mathbf{z}_{i}^{(m)},\mathbf{z})/\tau\right)
},
\end{equation}
where $\mathrm{sim}(\cdot,\cdot)$ denotes dot-product similarity and $\tau$ is a temperature parameter.
This objective pulls together embeddings from the same model while separating those from different models.

\paragraph{Model Center Construction.}
After training, we represent each model $M_m \in \mathcal{M}$ by a center in the embedding space. The center $\mathbf{c}_m$ is computed by averaging embeddings over all prompts in $\mathcal{X}$:
\begin{equation}
\mathbf{c}_m = \frac{1}{N} \sum_{i=1}^{N} \mathbf{z}_{i}^{(m)}.
\end{equation}
Optionally, outlier embeddings with large distances to a preliminary center can be removed before averaging.
The resulting centers serve as compact prototypes for subsequent ownership verification.

\subsection{Inference and Verification}

During inference, we measure the similarity between a suspicious model $M_t$ and the source model $M_s$.
For efficiency, we randomly sample a small probe set $\mathcal{X}' \subset \mathcal{X}$ and query $M_t$ to generate responses.
Following the extraction pipeline, we compute token-level log probabilities under $\mathcal{M}=\{M_s\}\cup\mathcal{F}$ and map them into the embedding space using the trained encoder $f_{\theta}(\cdot)$.
To reduce variance from stochastic generation, we average the embeddings over the probe set to obtain a robust representation:
\[
\bar{\mathbf{z}}^{(t)} = \frac{1}{|\mathcal{X}'|} \sum_{x_i \in \mathcal{X}'} \mathbf{z}_{i}^{(t)}.
\]

We subsequently perform verification by estimating the relative likelihood that $M_t$ originates from the source model. This is formulated as a soft posterior over model centers, where the similarity between $M_t$ and the source model is normalized against its similarities to the reference models:
\begin{equation}
p_s =
\frac{\exp\!\left(\kappa\, \mathrm{sim}(\bar{\mathbf{z}}^{(t)}, \mathbf{c}_s)\right)}
{\sum_{m \in \mathcal{M}} \exp\!\left(\kappa\, \mathrm{sim}(\bar{\mathbf{z}}^{(t)}, \mathbf{c}_m)\right)},
\end{equation}
where $\mathrm{sim}(\cdot,\cdot)$ denotes cosine similarity and $\kappa$ is a scaling factor controlling the sharpness of the distribution.
Finally, we define the verification score as:
\begin{equation}
\mathrm{Score}(M_t \rightarrow M_s)
=
p_s \cdot \mathrm{sim}(\bar{\mathbf{z}}^{(t)}, \mathbf{c}_s).
\end{equation}
This metric combines two complementary perspectives: $p_s$ captures the \emph{relative discriminativeness} of the source model against the reference models, while the cosine term ensures \emph{absolute proximity} in the feature space.
Consequently, a high score provides strong evidence that $M_t$ is derived from $M_s$.

\input{5_experiment}

\section{Conclusion}

Existing black-box fingerprinting methods are often unreliable because stochastic sampling causes generated texts to vary across repeated queries, making textual signals unstable and inconsistent. To address this, we propose Token-Level Probability Fingerprinting (\texttt{TOPF}), which exploits the stable probabilistic structure underlying generated text. \texttt{TOPF} formulates ownership verification as a relative comparison task, assessing whether a suspicious model is relatively closer to the protected source model than to reference models. Extensive experiments show that \texttt{TOPF} delivers robust and effective verification performance approaching white-box methods. Future work will examine its scalability to larger model families and its robustness against adaptive model theft attacks.

\newpage
\subsection*{AI use statement}
In this work, AI tools were used solely for language polishing and spelling correction. The research ideas, experimental design, result analysis, and conclusions were developed independently by the authors. All AI-assisted edits were carefully reviewed by the authors, who take full responsibility for the final content of this paper.

\subsection*{Ethics statement}
This work can help protect the intellectual property of large language models by supporting black-box ownership verification and reducing the risk of model theft. It may help detect unauthorized deployment, API wrapping, or derivative use of protected models. However, misuse of verification results may cause harm. False positives could lead to incorrect accusations against independently developed models, and the method could also be used for competitive auditing of external systems. Therefore, \texttt{TOPF} should be used as supporting technical evidence rather than definitive proof.

\subsection*{Reproducibility statement}
To support reproducibility, the methodology section provides the definitions and implementation details of the proposed method, while the experimental section describes the datasets, models, baselines, and evaluation settings. Additional details on the encoder architecture, training procedure, and hyperparameters are provided in the appendix. The implementation code is included in the supplementary material.



\bibliography{reference}
\bibliographystyle{iclr2027_conference}

\newpage
\appendix
\section{Encoder Model Architecture}
\label{app:encoder}

The CNN-Transformer encoder is designed to process log-probability sequences from multiple LRMs to generate robust embeddings.
As summarized in Table~\ref{tab:structure}, the architecture follows a multi-stage approach: (1) localized feature extraction, (2) linear projection, and (3) global temporal modeling.

The initial stage employs a parallel ensemble of 1D CNNs. 
Each independent branch is dedicated to processing the log-probability sequence from either the source model or one of the reference models.
These feature extractors map the raw sequences of shape $[B, 1, T]$ through four sequential layers with kernel sizes of 5, 3, 3, and 3, respectively.
This localized processing transforms the one-dimensional probability fluctuations into a higher-dimensional feature space consisting of 64 channels, capturing fine-grained temporal patterns.

Following the feature extraction, the outputs from all model-specific CNNs are concatenated along the channel dimension.
This composite vector is projected via a linear layer into a unified embedding space of dimension $d_{model} = 256$.
To preserve the sequential order of the tokens, position encodings are added to the projected embeddings before passing them through a layer normalization module.

The temporal modeling is performed by Transformer blocks.
Each block utilizes a multi-head self-attention mechanism with 8 heads and a FFN with a hidden dimension of 1024.
This allows the model to capture long-range dependencies across the entire sequence of probabilities.
The final response embedding is derived by applying masked mean pooling across the temporal dimension of the Transformer output, ensuring that only valid token positions contribute to the representation.

\begin{table}[htbp]
\centering
\caption{Structure of the CNN-Transformer Encoder.}
\label{tab:structure}
\begin{tabular}{lc}
\hline
Component & Configuration \\ \hline
CNN Kernel Sizes & [5, 3, 3, 3] \\
CNN Output Channels & 64 per model \\
Embedding Dimension $d_{model}$ & 256 \\
Transformer Layers & 2 \\
Attention Heads & 8 \\
FFN Dimension & 1024 \\
Pooling Strategy & Masked Mean \\ \hline
\end{tabular}
\end{table}

\begin{table}[htbp]
\centering
\caption{Training hyperparameters for the encoder.}
\label{tab:encoder_train}
\begin{tabular}{lc}
\hline
Hyperparameter & Configuration \\ \hline
Prompt Pool Size & 500 \\
Datasets & GSM8K, HotpotQA, LogiQA, MBPP, XSum \\
Queries per Dataset & 100 \\
Responses per Model per Prompt & 3 \\
Number of Reference Models & 3 \\
Total Responses & 6,000 \\
Optimizer & AdamW \\
Learning Rate & $1\times10^{-4}$ \\
Batch Size & 128 \\
Training Epochs & 30 \\
Weight Decay & 0.01 \\
Temperature $\tau$ & 0.07 \\ \hline
\end{tabular}
\end{table}

\section{Encoder Training Hyperparameters}
\label{app:encoder_train}

The training prompt pool consists of 500 prompts, constructed by sampling 100 queries from each of five open-source datasets: GSM8K, HotpotQA, LogiQA, MBPP, and XSum.
The source model and each reference model generate three responses for every prompt.
Under the default setting with three reference models, this produces a total of 6,000 responses.
The resulting response sequences are converted into probability matrices, which serve as the input to the encoder.
During testing, we randomly sample 20 prompts from the prompt pool as queries.

As shown in Table~\ref{tab:encoder_train}, the encoder is trained with supervised contrastive learning using the AdamW optimizer.
We use a learning rate of $1\times10^{-4}$, a batch size of 128, and train the model for 30 epochs.
The weight decay is set to 0.01, and the temperature coefficient $\tau$ is set to 0.07.
The encoder is lightweight and does not require careful hyperparameter tuning.
Training is performed on a single H200 GPU and takes about 3 minutes.

\section{Experimental Results}

\begin{table}[ht]
\centering
\setlength{\tabcolsep}{3pt}
\caption{Comparison with additional fingerprinting methods on LRMs. LSF requires gray-box access to output distributions, whereas CoTSRF and DuFFin operate under black-box settings.}
\label{tab:additional_baselines}
\resizebox{0.8\linewidth}{!}{%
\begin{tabular}{@{}l
    C{1.2cm}C{1.2cm}C{1.2cm}C{1.2cm}
    C{1.2cm}C{1.2cm}C{1.2cm}C{1.2cm}@{}}
\toprule
\multirow{2}{*}{Method}
& \multicolumn{4}{c}{DS-R1-8B}
& \multicolumn{4}{c}{Qwen3-8B} \\
\cmidrule(lr){2-5}
\cmidrule(lr){6-9}
& AUC & pAUC & TPR@3\% & MD
& AUC & pAUC & TPR@3\% & MD \\
\midrule
LSF
& \textbf{0.951} & \textbf{0.972} & \textbf{0.944} & \textbf{1.997}
& 0.949 & 0.915 & 0.833 & 1.741 \\

CoTSRF
& 0.890 & 0.744 & 0.833 & 1.361
& 0.853 & 0.744 & 0.778 & 1.648 \\

DuFFin
& 0.852 & 0.715 & 0.722 & 1.206
& 0.819 & 0.694 & 0.685 & 1.493 \\

\texttt{TOPF}
& 0.946 & 0.962 & 0.926 & 1.873
& \textbf{0.961} & \textbf{0.943} & \textbf{0.889} & \textbf{1.840} \\
\bottomrule
\end{tabular}%
}
\end{table}

\paragraph{Additional Baselines.}
To complement the main evaluation, we compare \texttt{TOPF} with three additional fingerprinting methods:
(1) LSF~\citep{yang2024fingerprint}, a gray-box method that uses the source model's final linear layer as a fingerprint and measures its alignment with the output-logit space of the suspicious model;
(2) CoTSRF~\citep{ren2025cotsrf}, a black-box method that learns model-specific chain-of-thought representations from generated responses through contrastive learning; and
(3) DuFFin~\citep{yan2025duffin}, a black-box framework that jointly extracts trigger-pattern and knowledge-level fingerprints from model responses.
As shown in Table~\ref{tab:additional_baselines}, \texttt{TOPF} performs comparably to LSF despite operating under the strict black-box setting.
The reported LSF results assume access to full-vocabulary output distributions, while restricting access to the top-20 logits reduces its performance to near-random guessing.
Compared with CoTSRF and DuFFin, \texttt{TOPF} consistently achieves higher performance across all metrics, with pronounced gains in pAUC and TPR@3\%.
These results demonstrate that token-level probability structures provide more robust ownership signals than fingerprints derived directly from textual responses.

\begin{table}[htbp]
\centering
\setlength{\tabcolsep}{4pt}
\caption{Ablation of encoder architecture, Transformer depth, and sequence length on DS-R1-8B. The default setting uses a CNN followed by a 2-layer Transformer with a sequence length of 1024.}
\label{tab:encoder_ablation}
\resizebox{0.57\linewidth}{!}{%
\begin{tabular}{@{}lC{1.2cm}C{1.2cm}C{1.4cm}C{1.2cm}@{}}
\toprule
Setting & AUC & pAUC & TPR@3\% & MD \\
\midrule
CNN Only         & 0.900 & 0.915 & 0.889 & 1.803 \\
Transformer Only & 0.924 & 0.923 & 0.907 & 1.854 \\
\midrule
1 Layer          & 0.944 & 0.942 & 0.917 & 1.867 \\
4 Layers         & 0.946 & 0.972 & 0.926 & 1.877 \\
\midrule
Seq. 256         & 0.922 & 0.923 & 0.898 & 1.863 \\
Seq. 512         & 0.941 & 0.940 & 0.917 & 1.869 \\
Default          & 0.946 & 0.962 & 0.926 & 1.873 \\
\bottomrule
\end{tabular}%
}
\end{table}

\paragraph{Encoder Ablation.}
As shown in Table~\ref{tab:encoder_ablation}, we ablate the encoder architecture, Transformer depth, and sequence length.
Removing the Transformer or CNN reduces AUC from 0.946 to 0.900 and 0.924, respectively, demonstrating that local probability patterns captured by the CNN complement the long-range token dependencies modeled by the Transformer.
Increasing the Transformer depth beyond two layers provides only marginal overall improvement, indicating that a shallow encoder is sufficient. 
Longer probability traces consistently improve performance, with AUC increasing from 0.922 at a sequence length of 256 to 0.946 at 1024, since longer traces preserve more information from the reasoning trajectory. 
Overall, \texttt{TOPF} remains robust across these configurations and exhibits limited sensitivity to the encoder hyperparameters.

\begin{table}[htbp]
\centering
\setlength{\tabcolsep}{4pt}
\caption{Effect of different reference model sets on \texttt{TOPF}.}
\label{tab:reference_selection}
\resizebox{0.65\linewidth}{!}{%
\begin{tabular}{@{}lC{1.3cm}C{1.5cm}C{1.3cm}C{1.5cm}@{}}
\toprule
\multirow{2}{*}{Reference Set}
& \multicolumn{2}{c}{DS-R1-8B}
& \multicolumn{2}{c}{Qwen3-8B} \\
\cmidrule(lr){2-3}
\cmidrule(lr){4-5}
& AUC & TPR@3\% & AUC & TPR@3\% \\
\midrule
Ministral + Phi + Gemma
& 0.946 & 0.926 & 0.961 & 0.889 \\
Ministral + Phi + MiMo
& 0.949 & 0.944 & 0.959 & 0.917 \\
\bottomrule
\end{tabular}%
}
\end{table}

\paragraph{Reference Model Ablation.}
We further evaluate the robustness of \texttt{TOPF} under the three-reference setting. As shown in Table~\ref{tab:reference_selection}, replacing Gemma-3-12B-Instruct with MiMo-7B-RL~\citep{xiaomi2025mimo} yields comparable performance. These results demonstrate that \texttt{TOPF} is robust to the choice of reference models and that a small reference set is sufficient to establish a stable decision boundary.

\begin{table}[htbp]
\centering
\setlength{\tabcolsep}{4pt}
\caption{Query budget, time cost per verification, and performance on DS-R1-8B using a single H200 GPU. The query budget is reported as the number of prompts multiplied by the number of generations per prompt.}
\label{tab:efficiency}
\resizebox{0.65\linewidth}{!}{%
\begin{tabular}{@{}lC{2.2cm}C{1.7cm}C{1.3cm}C{1.5cm}@{}}
\toprule
Method & \makecell{Query Budget\\Prompt $\times$ Gen.}
& Time Cost & AUC & TPR@3\% \\
\midrule
LLMmap    & $8 \times 1$   & 1.81 min  & 0.924 & 0.759 \\
MET       & $25 \times 5$  & 4.35 min  & 0.695 & 0.333 \\
ZEROPRINT & $10 \times 20$ & 7.36 min  & 0.693 & 0.389 \\
SEF       & $200 \times 1$ & 11.02 min & 0.900 & 0.556 \\
\texttt{TOPF} & $20 \times 1$ & 2.89 min
& \textbf{0.946} & \textbf{0.926} \\
\bottomrule
\end{tabular}%
}
\end{table}

\paragraph{Efficiency Analysis.}
We evaluate the query budget, verification time, and performance of each method on a single H200 GPU. As shown in Table~\ref{tab:efficiency}, \texttt{TOPF} completes each verification in 2.89 minutes while achieving the highest AUC and TPR@3\%. Its runtime is lower than most baselines despite evaluating responses under multiple models, demonstrating a balance between efficiency and effectiveness. Since ownership verification is an offline procedure, the computational cost is acceptable, and verification reliability remains the primary consideration.

\begin{table}[ht]
\centering
\setlength{\tabcolsep}{4pt}
\caption{Trace-level margins on DS-R1-8B. Source and Independent serve as empirical reference points.}
\label{tab:trace_margins}
\resizebox{0.6\linewidth}{!}{%
\begin{tabular}{@{}lC{2.0cm}C{2.0cm}@{}}
\toprule
Perturbation & $\Delta\rho\uparrow$ & $\Delta P\uparrow$ \\
\midrule
Source (upper bound)        & 0.421  & 0.126  \\
RAG                         & 0.394  & 0.096  \\
Quantization                & 0.410  & 0.108  \\
Fine-tuning                 & 0.363  & 0.063  \\
Merging                     & 0.113  & $-$0.033 \\
Independent (lower bound)   & 0.087  & $-$0.045 \\
\bottomrule
\end{tabular}%
}
\end{table}

\paragraph{Robustness Analysis.}
To explain the behavior of \texttt{TOPF} under different perturbations, we conduct a post hoc analysis and introduce two trace-level margins. For each response from a suspicious model, we compute token-level log-probability traces under the suspicious model itself~$d$, the source model~$s$, and each reference model~$f_k$.
(1) The correlation margin $\Delta\rho=\rho(d,s)-\max_k \rho(d,f_k)$, where $\rho$ is the Pearson correlation, measures whether the trajectory shape aligns more closely with the source than with the nearest reference. (2) The probability margin $\Delta P=\bar{p}_s-\max_k \bar{p}_{f_k}$, where $\bar{p}$ is the mean per-token probability, measures whether the source recognizes the response more confidently than any reference. As shown in Table~\ref{tab:trace_margins}, RAG and quantization retain both margins close to the source, indicating that these perturbations preserve probability patterns. Fine-tuning reduces both margins moderately due to parameter updates, while model merging shows the decline toward the independent lower bound, as blending parameters from different models dilutes probability structures.

\begin{table}[htbp]
\centering
\setlength{\tabcolsep}{3pt}
\caption{Per-perturbation AUC on DS-R1-8B. Param-Independent aggregates SP, RAG, and InPara. Param-Altering aggregates Adapter, FT, Merge, and Quant.}
\label{tab:per_perturbation_auc}
\resizebox{0.73\linewidth}{!}{%
\begin{tabular}{@{}lC{1.4cm}C{1.4cm}C{1.4cm}C{1.8cm}C{1.4cm}@{}}
\toprule
Perturbation & LLMmap & MET & SEF & ZEROPRINT & \texttt{TOPF} \\
\midrule
SP       & 1.000 & 0.864 & 1.000 & 1.000 & 1.000 \\
RAG      & 1.000 & 1.000 & 1.000 & 1.000 & 1.000 \\
InPara   & 1.000 & 1.000 & 1.000 & 1.000 & 1.000 \\
Adapter  & 1.000 & 0.586 & 0.964 & 0.907 & 1.000 \\
FT       & 0.869 & 0.663 & 0.909 & 0.886 & 1.000 \\
Merge    & 0.057 & 0.957 & 0.200 & 0.171 & 0.114 \\
Quant    & 0.857 & 0.457 & 1.000 & 0.886 & 1.000 \\
\midrule
Param-Independent & 1.000 & 0.910 & 1.000 & 1.000 & \textbf{1.000} \\
Param-Altering    & 0.842 & 0.643 & 0.873 & 0.829 & \textbf{0.919} \\
\bottomrule
\end{tabular}%
}
\end{table}

\paragraph{Per-Perturbation Verification Performance.}
We compare \texttt{TOPF} with black-box baselines on each perturbation type on DS-R1-8B. As shown in Table~\ref{tab:per_perturbation_auc}, all methods achieve high AUC on parameter-independent perturbations. For parameter-altering, \texttt{TOPF} attains the highest aggregate AUC of 0.919, achieving the best scores on Adapter, Fine-tuning, and Quantization. Model merging remains challenging for most methods, consistent with the trace-level analysis showing weakened alignment between merged and source models.

\begin{figure}[!htbp]
\centering
\includegraphics[width=0.47\linewidth]{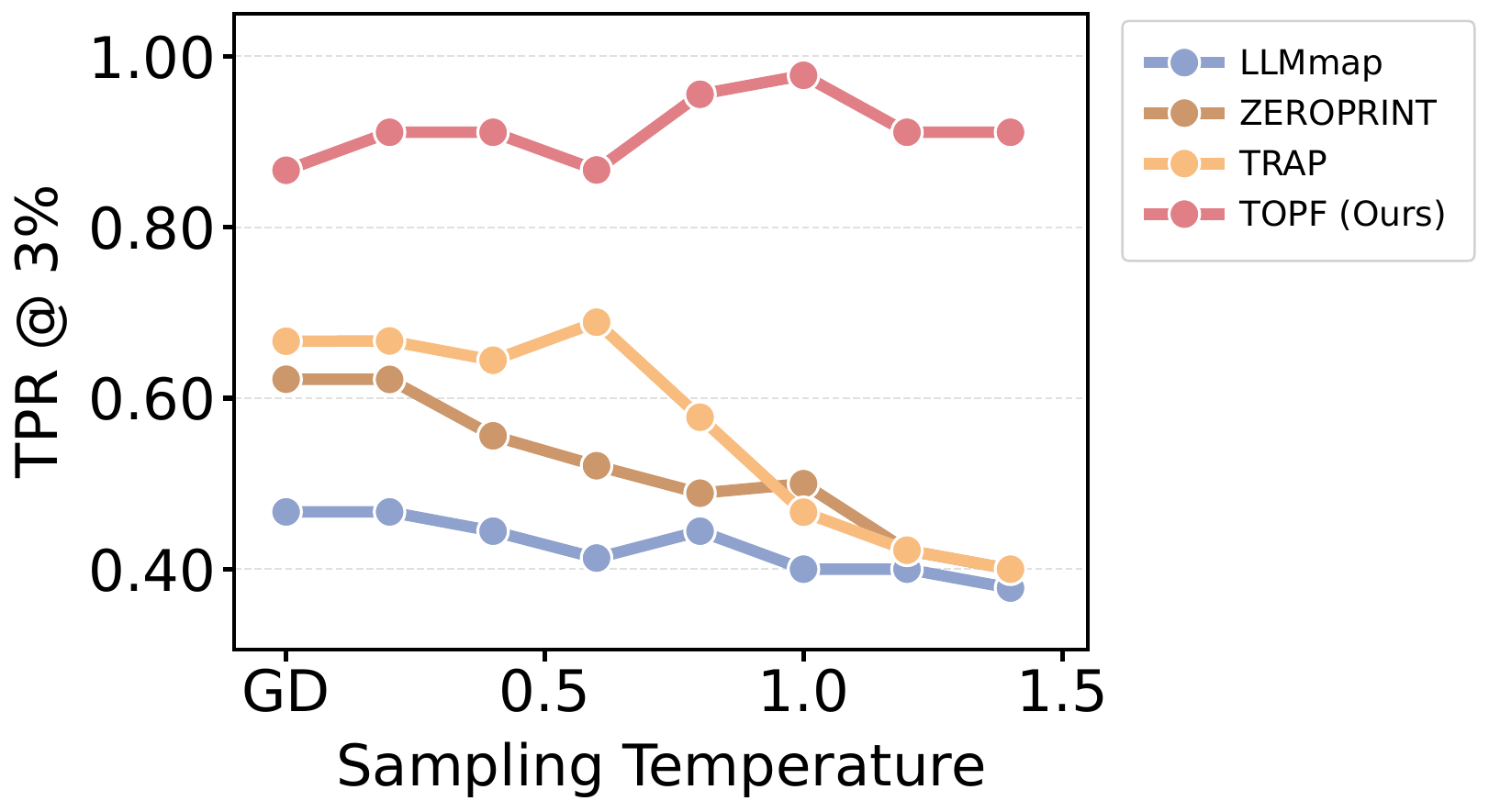} 
\caption{Performance across sampling temperatures on Mistral-7B. GD denotes greedy decoding.}
\label{fig:temp_mistral}
\end{figure}

\begin{table}[!htbp]
\centering
\setlength{\tabcolsep}{3pt}
\caption{Impact of the number of reference models on \texttt{TOPF}.}
\label{tab:num_models_all}
\resizebox{\linewidth}{!}{%
\begin{tabular}{@{}C{1cm}C{1.2cm}C{1.2cm}C{1.2cm}C{1.2cm} C{1.2cm}C{1.2cm}C{1.2cm}C{1.2cm} C{1.2cm}C{1.2cm}C{1.2cm}C{1.2cm}@{}}
\toprule
\multirow{2}{*}{Num.} & \multicolumn{4}{c}{DS-R1-8B} & \multicolumn{4}{c}{Qwen3-8B} & \multicolumn{4}{c}{GPT-OSS-20B} \\

\cmidrule(lr){2-5} 
\cmidrule(lr){6-9} 
\cmidrule(lr){10-13}

 & AUC & pAUC & TPR@3\% & MD & AUC & pAUC & TPR@3\% & MD & AUC & pAUC & TPR@3\% & MD \\ \midrule
1 & 0.980 & 0.972 & 0.944 & 1.981 & 0.996 & 0.980 & 0.981 & 1.996 & 0.937 & 0.937 & 0.922 & 1.878 \\
2 & 0.946 & 0.972 & 0.944 & 1.978 & 0.939 & 0.905 & 0.815 & 1.757 & 0.988 & 0.936 & 0.941 & 1.914 \\
3 & 0.946 & 0.962 & 0.926 & 1.873 & 0.961 & 0.943 & 0.889 & 1.840 & 0.987 & 0.950 & 0.902 & 1.901 \\ \bottomrule
\end{tabular}%
}
\end{table}

\begin{table}[!htbp]
\centering
\setlength{\tabcolsep}{3pt}
\caption{Impact of domain-specific probes on \texttt{TOPF}.}
\label{tab:probe_type_all}
\resizebox{\linewidth}{!}{%
\begin{tabular}{@{}lC{1.2cm}C{1.2cm}C{1.2cm}C{1.2cm} C{1.2cm}C{1.2cm}C{1.2cm}C{1.2cm} C{1.2cm}C{1.2cm}C{1.2cm}C{1.2cm}@{}}
\toprule
\multirow{2}{*}{Type} & \multicolumn{4}{c}{DS-R1-8B} & \multicolumn{4}{c}{Qwen3-8B} & \multicolumn{4}{c}{GPT-OSS-20B} \\

\cmidrule(lr){2-5} 
\cmidrule(lr){6-9} 
\cmidrule(lr){10-13}

& AUC & pAUC & TPR@3\% & MD & AUC & pAUC & TPR@3\% & MD & AUC & pAUC & TPR@3\% & MD \\ \midrule
GSM8K & 0.950 & 0.972 & 0.944 & 1.992 & 0.962 & 0.953 & 0.907 & 1.887 & 0.992 & 0.979 & 0.980 & 1.953 \\
HotpotQA & 0.946 & 0.962 & 0.926 & 1.938 & 0.937 & 0.896 & 0.796 & 1.726 & 0.978 & 0.950 & 0.902 & 1.911 \\
LogiQA & 0.950 & 0.972 & 0.944 & 1.970 & 0.881 & 0.886 & 0.778 & 1.663 & 0.975 & 0.898 & 0.824 & 1.787 \\
MBPP & 0.948 & 0.943 & 0.889 & 1.869 & 0.933 & 0.924 & 0.852 & 1.788 & 0.985 & 0.970 & 0.941 & 1.978 \\
XSum & 0.942 & 0.915 & 0.833 & 1.828 & 0.985 & 0.953 & 0.907 & 1.900 & 0.930 & 0.847 & 0.745 & 1.658 \\
All & 0.946 & 0.962 & 0.926 & 1.873 & 0.961 & 0.943 & 0.889 & 1.840 & 0.987 & 0.950 & 0.902 & 1.901 \\ \bottomrule
\end{tabular}%
}
\end{table}

\paragraph{Additional Experimental Results.}
We provide additional results that complement the experiments in the main text. Figure~\ref{fig:temp_mistral} reports performance across sampling temperatures on Mistral-7B.
Tables~\ref{tab:num_models_all} and~\ref{tab:probe_type_all} present the complete results for different numbers of reference models and domain-specific probes across AUC, pAUC, TPR@3\%, and MD.

\section{Limitations}
\label{app:limitation}
Although \texttt{TOPF} uses natural prompts from diverse domains, the selection of verification queries is not fully optimized in this work.
We mainly rely on a fixed prompt pool and randomly sampled probe sets, which may not always identify the most informative queries.
As a result, some probes may provide redundant or weak token-level probability traces, particularly under very limited query budgets.
Future work could investigate adaptive query selection strategies that choose more discriminative probes while further reducing the cost of black-box verification.


\section{Suspicious Models}
\label{app:suspicious}

\begin{longtable}{@{} l l >{\ttfamily}p{0.70\linewidth} @{}}
\caption{Suspicious models grouped by base model in the LRM setting.}
\label{tab:suspicious_lrm}\\
\toprule
Base Model & Type & model\_path \\
\midrule
\endfirsthead

\toprule
Base Model & Type & model\_path \\
\midrule
\endhead

\midrule
\multicolumn{3}{r}{\footnotesize Continued on next page.} \\
\midrule
\endfoot

\bottomrule
\endlastfoot

\multirow{15}{*}{DS-R1-8B}
& \multirow{5}{*}{Adapter}
& ai-medical/fine\_tuned\_deepseek\_v1\_empathy \\
& & sajjadhadi/Disease-Diagnosis-DeepSeek-R1- \\
& & Distill-Llama-8B \\
& & Seanie-lee/R1-8B-a0.0-thinksafe-r1-8B-v2 \\
& & Vijayendra/DeepSeek-Llama3.1-8B-DeepThinker-v1 \\
\cmidrule(l){2-3}
& \multirow{7}{*}{Fine-tuning}
& azzzacs/LogicCoder-8B \\
& & BlueBeck/LlamaAligned-DeepSeekR1-Distill-8b \\
& & enkryptai/DeepSeek-R1-Distill-Llama-8B \\
& & -Enkrypt-Aligned \\
& & huihui-ai/DeepSeek-R1-Distill-Llama-8B  \\
& & -abliterated \\
& & kangdawei/MMR-DAPO-8B \\
\cmidrule(l){2-3}
& Merge
& ForSureTesterSim/DeepLlama-Chimera-NSFW-8B \\
\cmidrule(l){2-3}
& \multirow{2}{*}{Quantization}
& unsloth/DeepSeek-R1-Distill-Llama-8B \\
& & -unsloth-bnb-4bit \\

\midrule
\multirow{12}{*}{Qwen3-8B}
& \multirow{4}{*}{Adapter}
& adamkarvonen/misaligned\_2\_qwen3-8B \\
& & mtzig/qwen3-8b-tfdark-lora2 \\
& & pkhare/qwen3-8b-biomedical \\
& & Seanie-lee/qwen3-8B-a0.0-thinksafe-8B \\
\cmidrule(l){2-3}
& \multirow{6}{*}{Fine-tuning}
& Goekdeniz-Guelmez/Josiefied-Qwen3-8B \\
& & -abliterated-v1 \\
& & hkust-nlp/WebExplorer-8B \\
& & marketeam/Qwen-Marketing \\
& & nvidia/Nemotron-Orchestrator-8B \\
& & Qwen/Qwen3Guard-Gen-8B \\
\cmidrule(l){2-3}
& Merge
& KaraKaraWitch/CavesOfQwen3-8b \\
\cmidrule(l){2-3}
& Quantization
& unsloth/Qwen3-8B-unsloth-bnb-4bit \\

\midrule
\multirow{10}{*}{GPT-OSS-20B}
& \multirow{4}{*}{Adapter}
& agency42/Helios-GPT-OSS-20B \\
& & Wildstash/dental-gpt-qlora \\
& & vrashad/gpt-oss-20b-medical-az \\
& & ZeroAgency/gpt-oss-20b-multilingual-reasoner \\
\cmidrule(l){2-3}
& \multirow{5}{*}{Fine-tuning}
& ArliAI/gpt-oss-20b-Derestricted \\
& & aoxo/gpt-oss-20b-uncensored \\
& & openai/gpt-oss-safeguard-20b \\
& & ValiantLabs/gpt-oss-20b-Esper3.1 \\
& & ValiantLabs/gpt-oss-20b-ShiningValiant3 \\
\cmidrule(l){2-3}
& Quantization
& unsloth/gpt-oss-20b-unsloth-bnb-4bit \\

\end{longtable}

\newpage

\begin{longtable}{@{} l l >{\ttfamily}p{0.73\linewidth} @{}}
\caption{Suspicious models grouped by base model in the LLM setting.}
\label{tab:suspicious_llm}\\
\toprule
Base Model & Type & model\_path \\
\midrule
\endfirsthead

\toprule
Base Model & Type & model\_path \\
\midrule
\endhead

\midrule
\multicolumn{3}{r}{\footnotesize Continued on next page.} \\
\midrule
\endfoot

\bottomrule
\endlastfoot

\multirow{11}{*}{Llama3.1-8B}
& \multirow{3}{*}{Adapter}
& LlamaFactoryAI/Llama-3.1-8B-Instruct-cv-\\
& & job-description-matching \\
& & chchen/Llama-3.1-8B-Instruct-PsyCourse-fold7 \\
\cmidrule(l){2-3}
& \multirow{4}{*}{Fine-tuning}
& RedHatAI/Llama-3.1-8B-tldr \\
& & ValiantLabs/Llama3.1-8B-Fireplace2 \\
& & mlabonne/Meta-Llama-3.1-8B-Instruct-abliterated \\
& & proxectonos/Llama-3.1-Carballo \\
\cmidrule(l){2-3}
& \multirow{3}{*}{Merge}
& Xiaojian9992024/Llama3.1-8B-ExtraMix \\
& & gaverfraxz/Meta-Llama-3.1-8B-Instruct- \\
& & HalfAbliterated-TIES \\
\cmidrule(l){2-3}
& Quantization
& unsloth/Llama-3.1-8B-Instruct-unsloth-bnb-4bit \\

\midrule
\multirow{9}{*}{Qwen2.5-7B}
& \multirow{2}{*}{Adapter}
& SeeFlock/task-12-Qwen-Qwen2.5-7B-Instruct \\
& & fangcaotank/task-10-Qwen-Qwen2.5-7B-Instruct \\
\cmidrule(l){2-3}
& \multirow{4}{*}{Fine-tuning}
& Qwen/Qwen2.5-Coder-7B-Instruct \\
& & Qwen/Qwen2.5-Math-7B \\
& & WangCa/Qwen2.5-7B-Medicine \\
& & huihui-ai/Qwen2.5-7B-Instruct-abliterated-v2 \\
\cmidrule(l){2-3}
& \multirow{2}{*}{Merge}
& Locutusque/StockQwen-2.5-7B \\
& & bunnycore/QevaCoT-7B-Stock \\
\cmidrule(l){2-3}
& Quantization
& unsloth/Qwen2.5-7B-Instruct-unsloth-bnb-4bit \\

\midrule
\multirow{8}{*}{Mistral-7B}
& \multirow{2}{*}{Adapter}
& chaymaemerhrioui/mistral-Brain\_Model\_ACC\_Trainer \\
& & tomdickson/riksbanken-mistral-lora \\
\cmidrule(l){2-3}
& \multirow{4}{*}{Fine-tuning}
& KurmaAI/AQUA-7B \\
& & evolveon/Mistral-7B-Instruct-v0.3-abliterated \\
& & openfoodfacts/spellcheck-mistral-7b \\
& & shenzhi-wang/Mistral-7B-v0.3-Chinese-Chat \\
\cmidrule(l){2-3}
& Merge
& grimjim/Mistral-7B-Instruct-demi-merge-v0.3-7B \\
\cmidrule(l){2-3}
& Quantization
& unsloth/mistral-7b-instruct-v0.3-bnb-4bit \\

\end{longtable}

\newpage
\section{Similarity Matrices}
\label{app:similarity}

\begin{figure}[ht]
\centering
    \begin{subfigure}{0.48\textwidth}
        \centering
        \includegraphics[height=0.8\textheight, keepaspectratio]{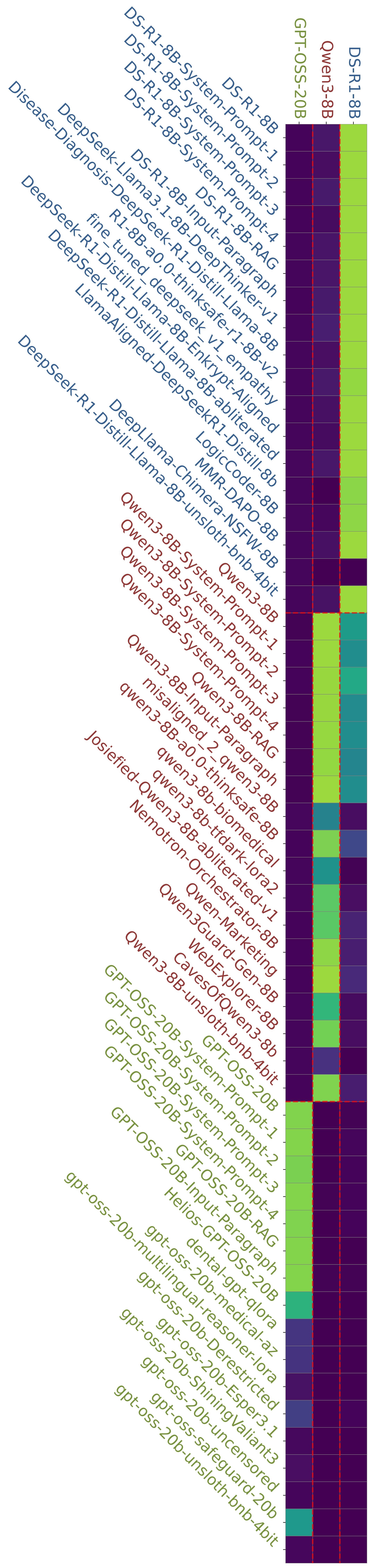}
        \caption{Results on LRMs.}
    \end{subfigure}
    \hfill
    \begin{subfigure}{0.48\textwidth}
        \centering
        \includegraphics[height=0.8\textheight, keepaspectratio]{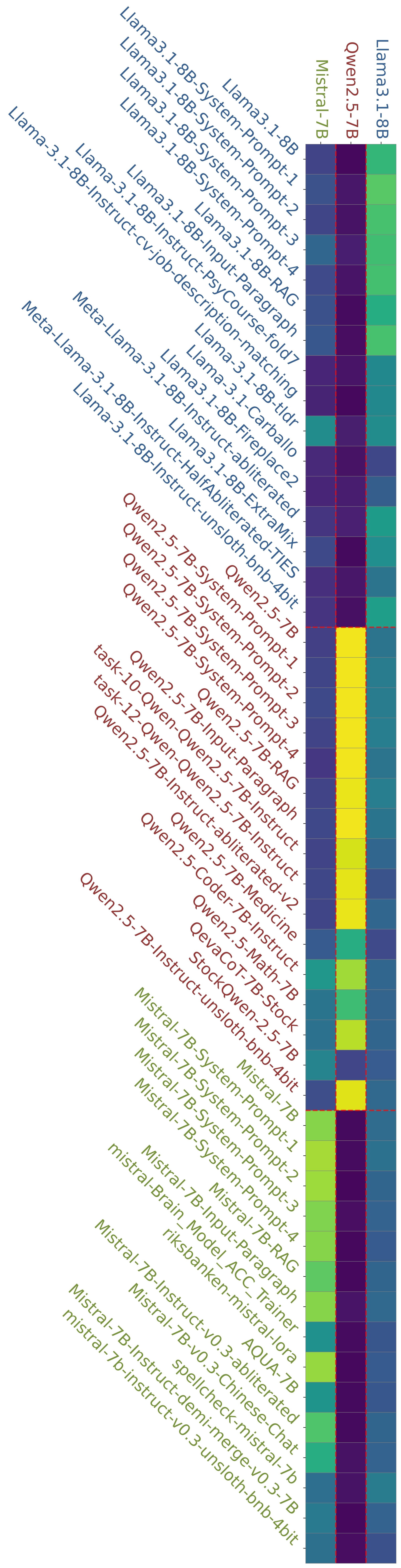}
        \caption{Results on standard LLMs.}
    \end{subfigure}

\caption{Complete similarity matrices for \texttt{TOPF} results.}
\label{fig:similarity_all}
\end{figure}

\end{document}

%% file: 5_experiment.tex
\begin{table}[t]
\centering
\setlength{\tabcolsep}{1pt}
\caption{Comparison of intrinsic fingerprinting methods on both LRMs and standard LLMs.
White-box and black-box methods are indicated by \whitebox\ and \blackbox, respectively.
The table reports $\uparrow$ AUC, $\uparrow$ pAUC, $\uparrow$ TPR@3\%, and $\uparrow$ MD.
The results of the best black-box methods are highlighted in bold.
Notably, \texttt{TOPF} achieves performance comparable to white-box methods across models.
}
\label{tab:main}

\begin{subtable}{\linewidth}
\centering
\caption{Results on LRMs.
}
\resizebox{\linewidth}{!}{%
\begin{tabular}{@{}l
    C{1.2cm}C{1.2cm}C{1.2cm}C{1.2cm}
    C{1.2cm}C{1.2cm}C{1.2cm}C{1.2cm}
    C{1.2cm}C{1.2cm}C{1.2cm}C{1.2cm}@{}}
\toprule
\multirow{2}{*}{Method} & 
\multicolumn{4}{c}{DS-R1-8B} &
\multicolumn{4}{c}{Qwen3-8B} &
\multicolumn{4}{c}{GPT-OSS-20B} \\

\cmidrule(lr){2-5} \cmidrule(lr){6-9} \cmidrule(lr){10-13}

 &
AUC & pAUC & TPR@3\% & MD &
AUC & pAUC & TPR@3\% & MD &
AUC & pAUC & TPR@3\% & MD \\

\midrule

HuRef \whitebox
 & 0.917 & 0.957 & 0.917 & 2.073
 & 0.960 & 0.957 & 0.917 & 1.942
 & 0.958 & 0.953 & 0.909 & 1.978 \\

PDF \whitebox
 & 0.841 & 0.915 & 0.833 & 1.937
 & 0.922 & 0.957 & 0.917 & 1.973
 & 0.915 & 0.953 & 0.909 & 2.014 \\

REEF \whitebox
 & 0.986 & 0.957 & 0.917 & 1.878
 & 0.960 & 0.957 & 0.917 & 1.870
 & 0.937 & 0.868 & 0.727 & 1.702 \\

\midrule

LLMmap \blackbox
 & 0.924 & 0.877 & 0.759 & 1.188
 & 0.602 & 0.687 & 0.389 & 0.595
 & 0.818 & 0.732 & 0.510 & 0.907 \\

MET \blackbox
 & 0.695 & 0.665 & 0.333 & 1.059
 & 0.676 & 0.676 & 0.352 & 1.074
 & 0.686 & 0.686 & 0.373 & 1.135 \\

SEF \blackbox
 & 0.900 & 0.772 & 0.556 & 1.359
 & 0.563 & 0.687 & 0.389 & 0.549
 & 0.667 & 0.759 & 0.529 & 0.807 \\

ZEROPRINT \blackbox
 & 0.693 & 0.687 & 0.389 & 1.186
 & 0.700 & 0.691 & 0.407 & 1.174
 & 0.702 & 0.698 & 0.412 & 1.153 \\

NLL \blackbox
 & \textbf{0.965} & 0.727 & 0.722 & 1.730
 & 0.957 & 0.647 & 0.630 & 1.773
 & 0.899 & 0.838 & 0.706 & 1.567 \\

\texttt{TOPF} (Ours) \blackbox
 & 0.946 & \textbf{0.962} & \textbf{0.926} & \textbf{1.873}
 & \textbf{0.961} & \textbf{0.943} & \textbf{0.889} & \textbf{1.840}
 & \textbf{0.987} & \textbf{0.950} & \textbf{0.902} & \textbf{1.901} \\

\bottomrule
\end{tabular}%
}
\end{subtable}

\vspace{0.5em}

\begin{subtable}{\linewidth}
\centering
\caption{Results on standard LLMs.
}
\resizebox{\linewidth}{!}{%
\begin{tabular}{@{}l
    C{1.2cm}C{1.2cm}C{1.2cm}C{1.2cm}
    C{1.2cm}C{1.2cm}C{1.2cm}C{1.2cm}
    C{1.2cm}C{1.2cm}C{1.2cm}C{1.2cm}@{}}
\toprule
\multirow{2}{*}{Method} & 
\multicolumn{4}{c}{Llama3.1-8B} &
\multicolumn{4}{c}{Qwen2.5-7B} &
\multicolumn{4}{c}{Mistral-7B} \\

\cmidrule(lr){2-5} \cmidrule(lr){6-9} \cmidrule(lr){10-13}

 &
AUC & pAUC & TPR@3\% & MD &
AUC & pAUC & TPR@3\% & MD &
AUC & pAUC & TPR@3\% & MD \\

\midrule

HuRef \whitebox
 & 0.903 & 0.950 & 0.903 & 2.056
 & 0.908 & 0.950 & 0.903 & 1.748
 & 0.903 & 0.945 & 0.893 & 1.971 \\

PDF \whitebox
 & 0.906 & 0.950 & 0.903 & 1.772
 & 0.954 & 0.950 & 0.903 & 1.876
 & 0.895 & 0.945 & 0.893 & 1.865 \\

REEF \whitebox
 & 0.954 & 0.950 & 0.903 & 1.891
 & 0.742 & 0.736 & 0.485 & 1.085
 & 0.915 & 0.945 & 0.893 & 1.768 \\

\midrule

LLMmap \blackbox
 & 0.507 & 0.697 & 0.409 & 0.152
 & 0.692 & 0.707 & 0.429 & 0.731
 & 0.673 & 0.699 & 0.413 & 0.623 \\

MET \blackbox
 & 0.714 & 0.714 & 0.429 & 1.202
 & 0.684 & 0.684 & 0.368 & 1.070
 & 0.696 & 0.696 & 0.392 & 1.171 \\

SEF \blackbox
 & 0.418 & 0.676 & 0.368 & 0.977
 & 0.637 & 0.676 & 0.368 & 0.492
 & 0.713 & 0.704 & 0.435 & 0.766 \\

ZEROPRINT \blackbox
 & 0.685 & 0.707 & 0.429 & 1.134
 & 0.832 & 0.715 & 0.429 & 1.234
 & 0.821 & 0.759 & 0.521 & 1.303 \\

TRAP \blackbox
 & \textbf{0.875} & \textbf{0.867} & \textbf{0.737} & \textbf{1.501}
 & 0.779 & 0.763 & 0.507 & 1.295
 & 0.875 & 0.844 & 0.693 & 1.578 \\

\texttt{TOPF} (Ours) \blackbox
 & 0.822 & 0.780 & 0.542 & 1.389
 & \textbf{1.000} & \textbf{1.000} & \textbf{1.000} & \textbf{2.037}
 & \textbf{0.985} & \textbf{0.940} & \textbf{0.867} & \textbf{1.857} \\

\bottomrule
\end{tabular}%
}
\end{subtable}

\end{table}

\section{Experiments}

\subsection{Experimental Setup}

\paragraph{Source Models.}
We consider both large reasoning models (LRMs) and standard LLMs.
For LRMs, we use DeepSeek-R1-Distill-Llama-8B (DS-R1-8B)~\citep{guo2025deepseek}, Qwen3-8B~\citep{yang2025qwen3}, and GPT-OSS-20B~\citep{agarwal2025gptoss}.
For LLMs, we use Llama-3.1-8B-Instruct (Llama3.1-8B)~\citep{grattafiori2024llama}, Qwen2.5-7B-Instruct (Qwen2.5-7B)~\citep{yang2025qwen25}, and Mistral-7B-Instruct (Mistral-7B)~\citep{jiang2023mistral7b}.

\paragraph{Suspicious Models.}
To evaluate practical robustness, we construct suspicious models using diverse post-development strategies.
We consider parameter-independent deployments, including (1) different system prompts, (2) retrieval-augmented generation (RAG), and (3) input paraphrasing.
We further include parameter-altering transformations, such as (4) fine-tuning, (5) model merging, and (6) quantization.
The sampling temperature is set to 0.6.
Details are provided in Appendix~\ref{app:suspicious}.

\paragraph{Reference Models.}
To ensure a clear train-test separation, the reference models used for encoder training are kept disjoint from the suspicious models used for evaluation.
For LRMs, we use Ministral-3-8B-Reasoning~\citep{liu2026ministral3}, Gemma-3-12B-Instruct~\citep{kamath2025gemma3}, and Phi-4-Reasoning~\citep{abdin2025phi4}.
For LLMs, we adopt Apertus-8B-Instruct~\citep{cano2025apertus}, Gemma-3-12B-Instruct, and Olmo-3-7B-Instruct~\citep{ettinger2025olmo3}.

\paragraph{Metrics.}
We adopt four evaluation metrics.
(1) Area Under the ROC Curve (AUC).
(2) Partial AUC (pAUC) with false positive rate (FPR) restricted to $[0, 0.05]$, focusing on detection performance in low-FPR regimes. This metric is standardized to $[0, 1]$.
(3) TPR@3\%, which reports the TPR when the FPR is fixed at 3\%.
(4) Mahalanobis Distance (MD), measuring the statistical separability between source and suspicious model.

\paragraph{Baselines.}
We compare \texttt{TOPF} with representative white-box and black-box methods.
For white-box baselines, we compare:
(1) HuRef~\citep{zeng2024huref}, which constructs invariant fingerprints from Transformer weights;
(2) PDF~\citep{yoon2025intrinsic}, which captures parameter distribution statistics;
(3) REEF~\citep{zhang2025reef}, which measures representation similarity via centered kernel alignment.
For black-box baselines, we evaluate:
(4) LLMmap~\citep{pasquini2025llmmap}, which embeds model responses into learned feature spaces;
(5) MET~\citep{gao2025model}, which performs statistical hypothesis testing on output distributions;
(6) SEF~\citep{shao2025sok}, which extracts stylistic features from generated responses;
(7) TRAP~\citep{gubri2024trap}, which leverages targeted prompt–response pairs optimized via GCG~\citep{zou2023universal} on the source model;
(8) ZEROPRINT~\citep{shao2025reading}, which estimates gradient-based fingerprints through zeroth-order approximation.
We additionally construct
(9) NLL, a perplexity-based baseline that measures similarity using the inverse perplexity assigned by the source model to responses from the suspicious model.

\begin{figure}[t]
\centering

\begin{subfigure}{0.99\textwidth}
\centering
\includegraphics[width=\linewidth]{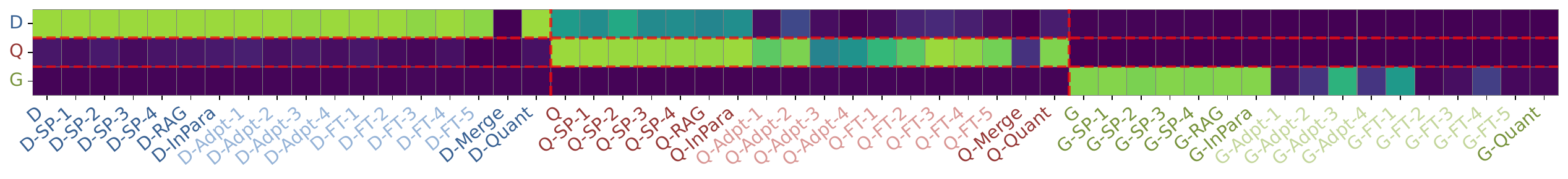}
\caption{LRMs: D = DS-R1-8B, Q = Qwen3-8B, G = GPT-OSS-20B.}
\end{subfigure}


\begin{subfigure}{0.94\textwidth}
\centering
\includegraphics[width=\linewidth]{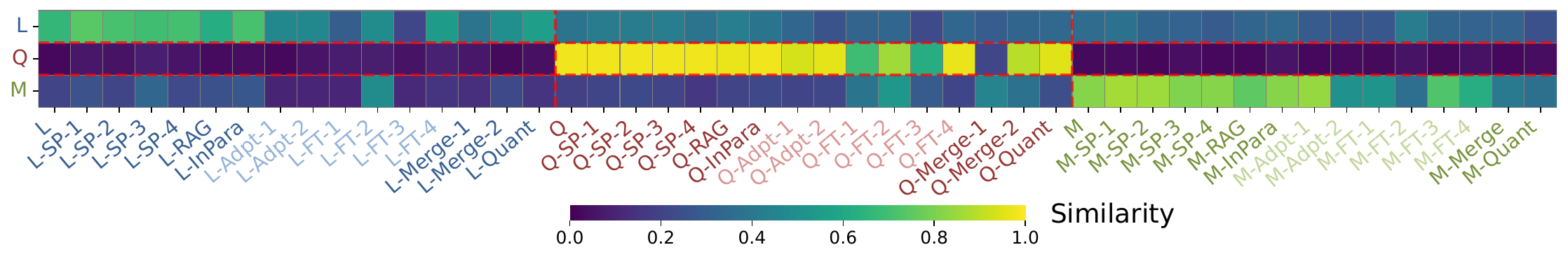}
\caption{LLMs: L = Llama3.1-8B, Q = Qwen2.5-7B, M = Mistral-7B.}
\end{subfigure}

\caption{Similarity matrices for \texttt{TOPF} results. Similarity (0–1) is computed between source models (row labels) and suspicious/derived variants (column labels). Derived variants include system prompt (SP), RAG deployment (RAG), input paraphrase (InPara), adapters (Adpt), fine-tuning (FT), model merging (Merge), and quantization (Quant).}
\label{fig:similarity}

\end{figure}

\subsection{Main Results}

\paragraph{Results on LRMs.}
Table~\ref{tab:main}(a) reports the results on LRMs.
We exclude TRAP because its GCG-based prompt optimization fails to converge on LRMs.
Existing black-box methods exhibit clear limitations: LLMmap, MET, SEF, and ZEROPRINT achieve low AUC and TPR@3\% across all models.
The perplexity-based NLL baseline attains higher AUC values (\eg, 0.965 on DS-R1-8B), but its pAUC and TPR@3\% remain below \texttt{TOPF}.
In contrast, \texttt{TOPF} achieves effective performance among black-box methods across all LRMs.
On DS-R1-8B, \texttt{TOPF} attains a pAUC of 0.962 and a TPR@3\% of 0.926, approaching the performance of white-box methods.
\texttt{TOPF} also yields the largest MD values among black-box approaches, indicating stronger statistical separation.
Overall, \texttt{TOPF} delivers stable and competitive verification performance while requiring only black-box access.

\paragraph{Results on LLMs.}


Table~\ref{tab:main}(b) reports the results on instruction-tuned LLMs. In this setting, NLL performs poorly in low-FPR regimes and is therefore omitted from comparison. \texttt{TOPF} leverages Chain-of-Thought prompting to induce explicit reasoning processes, enabling informative fingerprints. Other black-box baselines show limited effectiveness, with low TPR@3\% and MD values. On Qwen2.5-7B, \texttt{TOPF} achieves AUC, pAUC, and TPR@3\% scores of 1.000, outperforming both black-box and white-box baselines. On Llama3.1-8B, TRAP performs best, while \texttt{TOPF} ranks second with an AUC of 0.822 and a TPR@3\% of 0.542. This gap may be attributed to the more stable generation behavior of LLMs, which better aligns with TRAP's optimization assumption.

\begin{wraptable}[12]{r}{0.6\textwidth}
\centering
\setlength{\tabcolsep}{1pt}
\caption{Verification accuracy (\%) on different models.
}
\label{tab:acc}
\resizebox{\linewidth}{!}{%
\begin{tabular}{@{}lC{1.4cm}C{1.4cm}C{1.4cm}C{1.4cm}C{1.4cm}C{1.4cm}@{}}
\toprule
Method & DS-R1 & Qwen3 & GPT-OSS & Llama3.1 & Qwen2.5 & Mistral \\ \midrule
HuRef \whitebox & 87.40 & 87.71 & 84.20 & 72.75 & 83.25 & 88.50 \\
PDF \whitebox & 90.42 & 88.75 & 87.20 & 70.25 & 88.50 & 76.75 \\
REEF \whitebox & 91.88 & 87.71 & 77.80 & 85.00 & 87.50 & 82.75 \\ \midrule
LLMmap \blackbox & 85.39 & 46.45 & 76.35 & 40.62 & 53.03 & 58.18 \\
MET \blackbox & 34.21 & 34.21 & 32.43 & 33.33 & 55.91 & 58.18 \\
SEF \blackbox & 73.29 & 40.27 & 54.16 & 28.94 & 46.82 & 58.94 \\
ZEROPRINT \blackbox & 58.68 & 59.19 & 57.52 & 64.09 & 76.82 & 76.06 \\
\texttt{TOPF} (Ours) \blackbox & \textbf{95.39} & \textbf{93.92} & \textbf{95.80} & \textbf{81.67} & \textbf{89.09} & \textbf{89.60} \\ \bottomrule
\end{tabular}%
}
\end{wraptable}

\paragraph{Verification Accuracy.}
To calibrate the threshold in the open-world setting, we use a held-out validation set and keep the threshold fixed at test time.
Specifically, we compute the final verification scores for positive and negative validation pairs, and set the threshold as the midpoint between their mean scores.
A suspicious model is identified as derived only when its verification score exceeds this threshold.
We split the models into 70\% validation and 30\% test sets, and report averages over 20 random trials.
As shown in Table~\ref{tab:acc}, \texttt{TOPF} achieves strong accuracy across different source models, reaching 95.39\% on DS-R1-8B, 93.92\% on Qwen3-8B, and 95.80\% on GPT-OSS-20B.
Overall, \texttt{TOPF} consistently outperforms black-box baselines and achieves the best performance on most source models.


\subsection{Mechanism Analysis}

\paragraph{Similarity Matrices.}
Figure~\ref{fig:similarity} presents the similarity matrices produced by \texttt{TOPF} under both LRM and LLM settings. In both cases, models derived from the same source show high similarity, whereas non-derived models remain clearly separated. This pattern is consistent across parameter-independent deployments, including different system prompts, RAG, and input paraphrasing, indicating that \texttt{TOPF} captures probability patterns that are invariant to such changes. Quantized models also retain high similarity, suggesting that numerical precision has limited impact. In contrast, fine-tuning and model merging lead to lower similarity scores, likely due to shifts in model parameters and generation distributions. Model details are provided in Appendix~\ref{app:similarity}.




\begin{figure}[t]
\centering
\begin{subfigure}{0.47\textwidth}
    \centering
    \includegraphics[width=\linewidth]{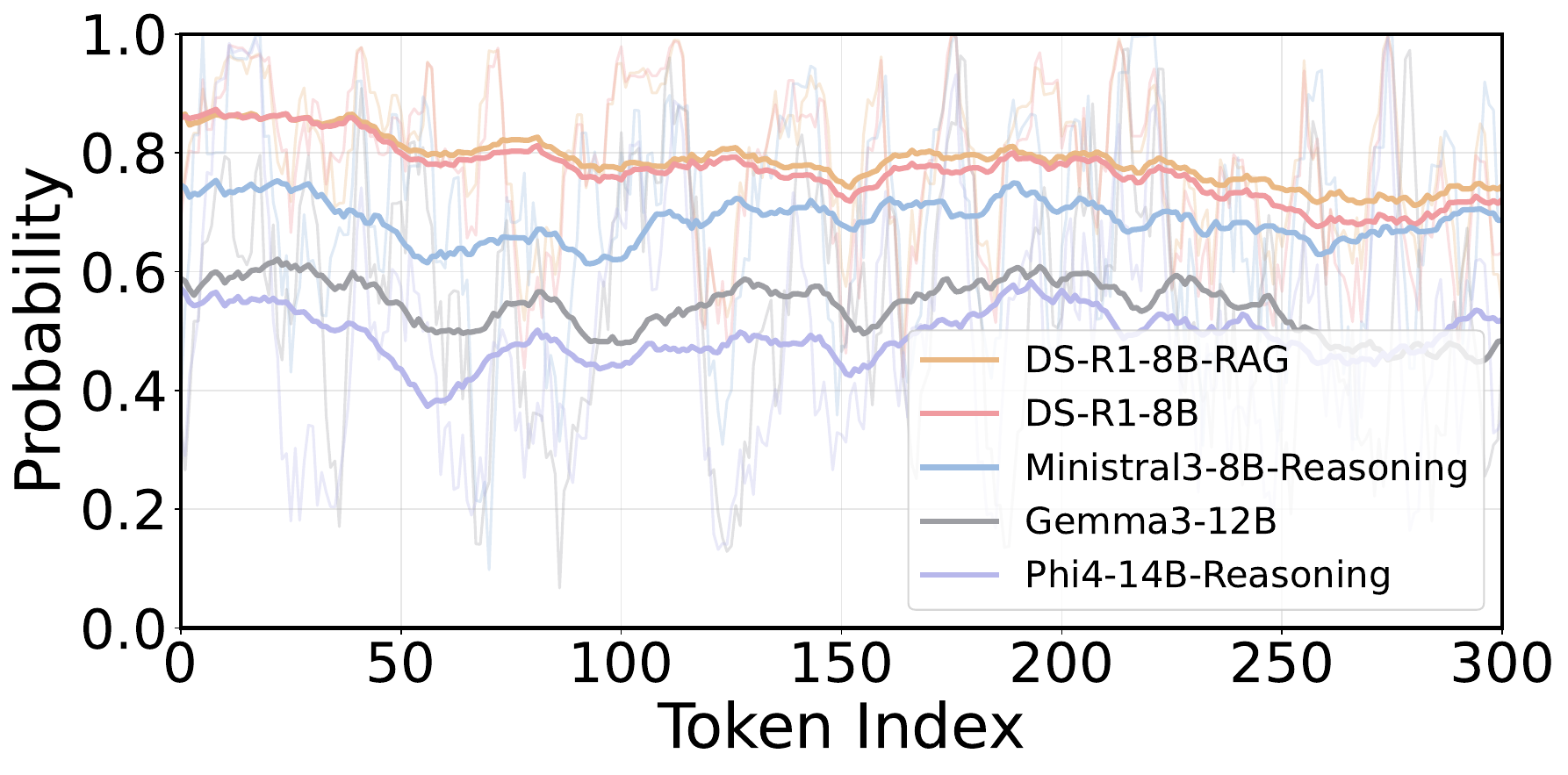}
    \caption{RAG derived model.}
\end{subfigure}
\hspace{0.03\textwidth}
\begin{subfigure}{0.47\textwidth}
    \centering
    \includegraphics[width=\linewidth]{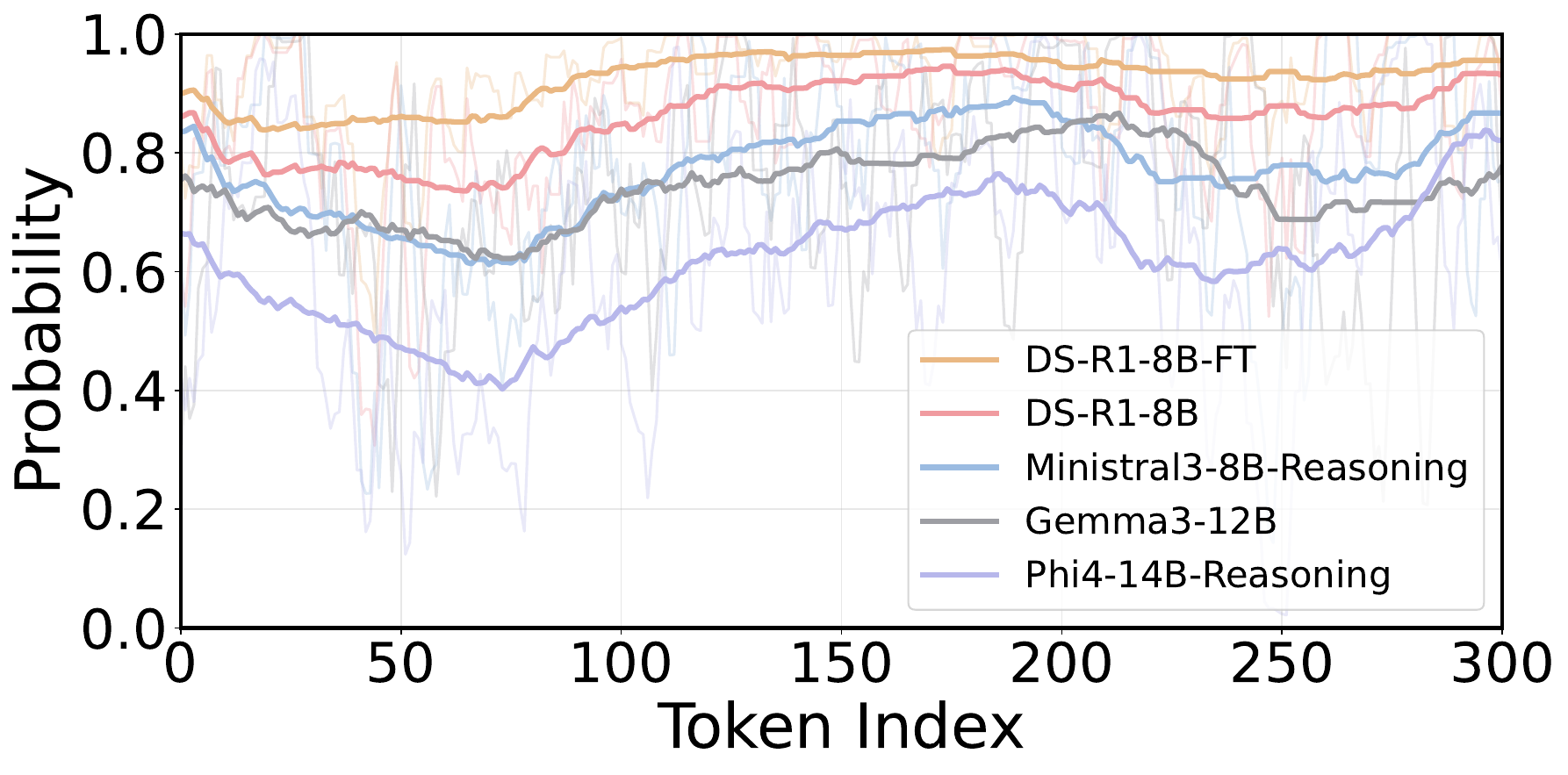}
    \caption{Fine-tuned derived model.}
\end{subfigure}


\begin{subfigure}{0.47\textwidth}
    \centering
    \includegraphics[width=\linewidth]{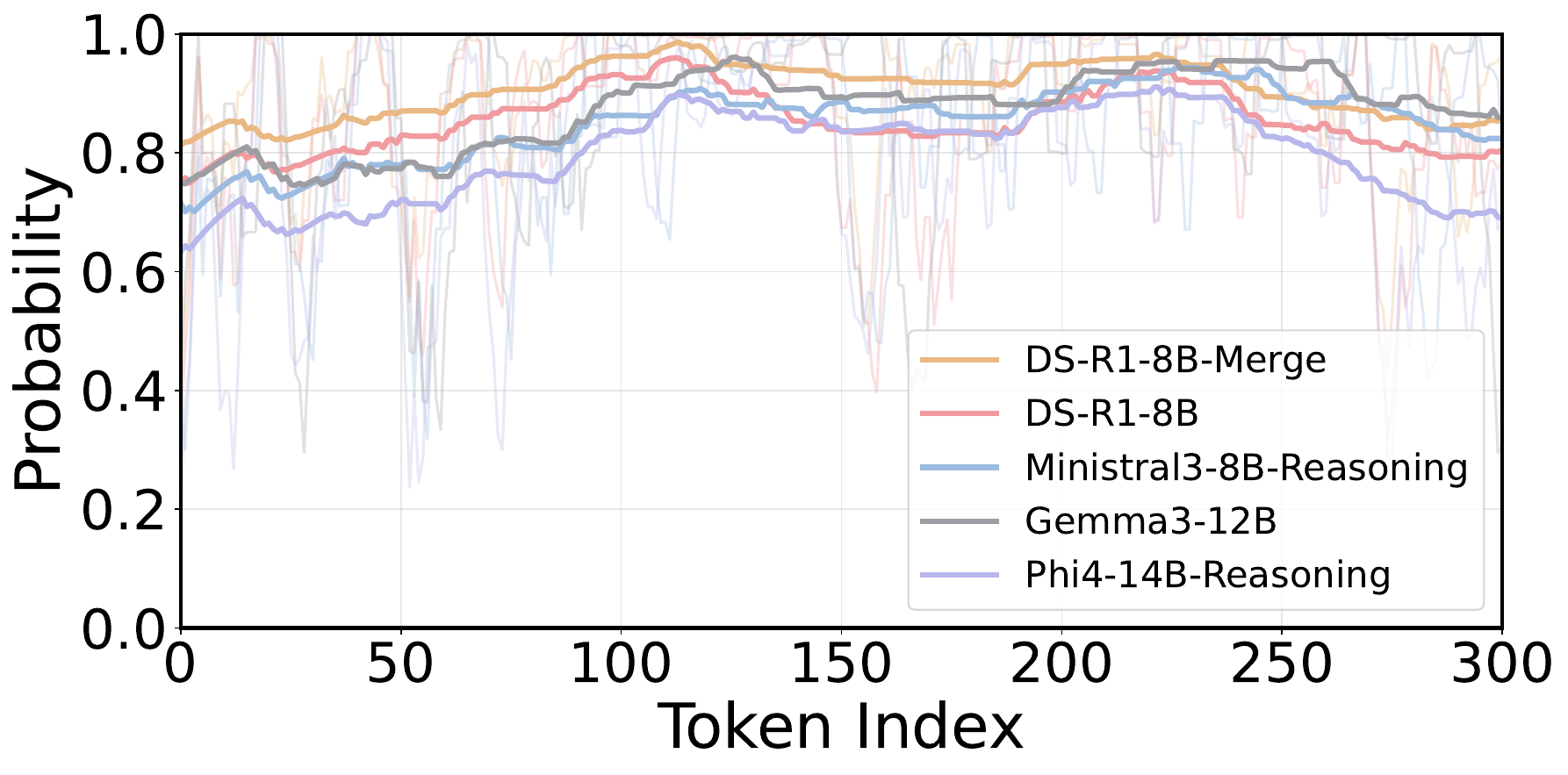}
    \caption{Merged derived model.}
\end{subfigure}
\hspace{0.03\textwidth}
\begin{subfigure}{0.47\textwidth}
    \centering
    \includegraphics[width=\linewidth]{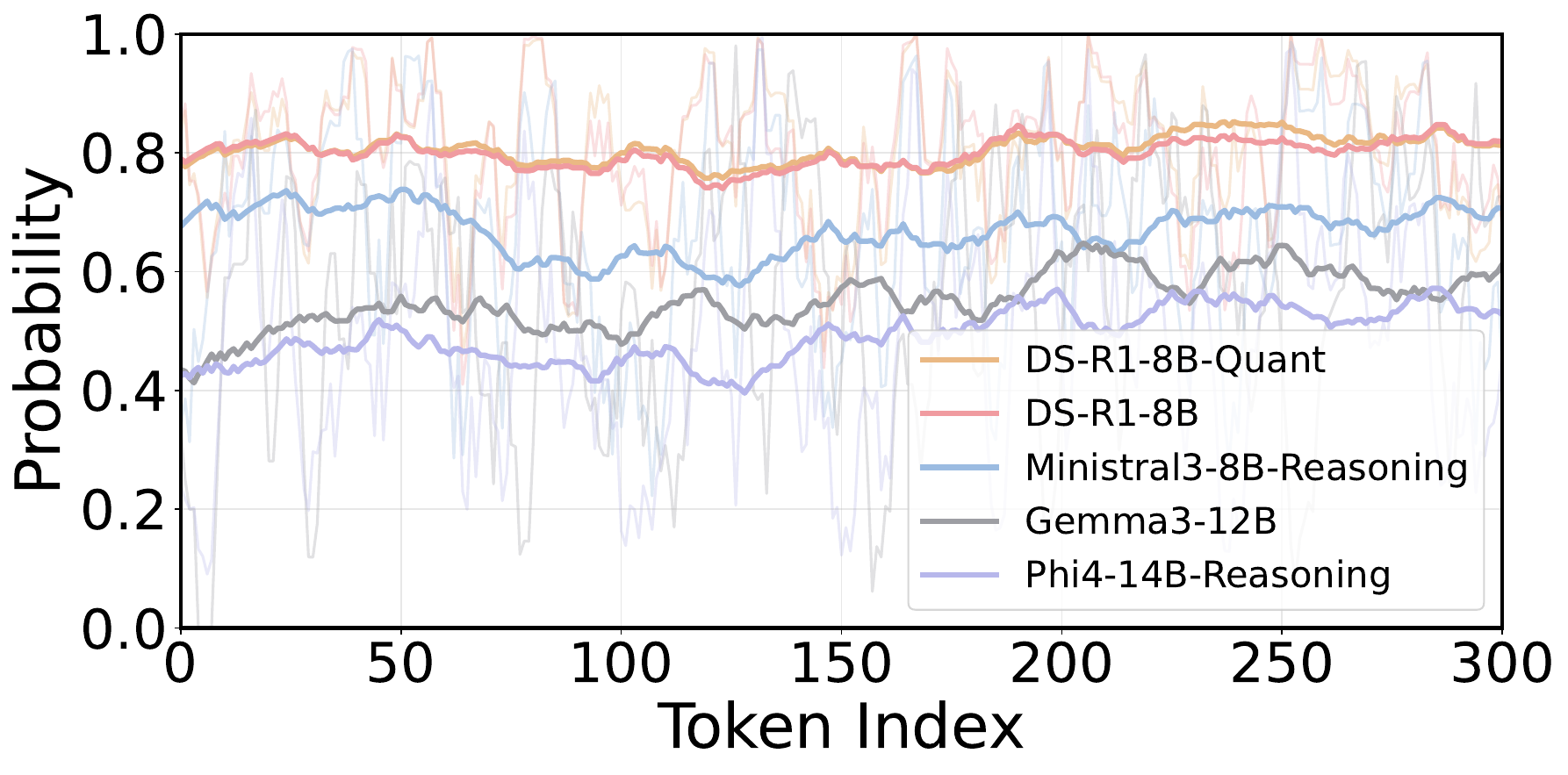}
    \caption{Quantized derived model.}
\end{subfigure}

\caption{Token-level probabilities for the response generated by DS-R1-8B-Derived under different models.
The derived models exhibit probability distributions that closely align with the source model, indicating similar generation behavior. Notably, the variants obtained via RAG and quantization are more similar to the original model than those obtained via fine-tuning and model merging.}
\label{fig:prob_all}
\end{figure}

\begin{wrapfigure}[15]{r}{0.5\textwidth}
\centering
\includegraphics[width=\linewidth]{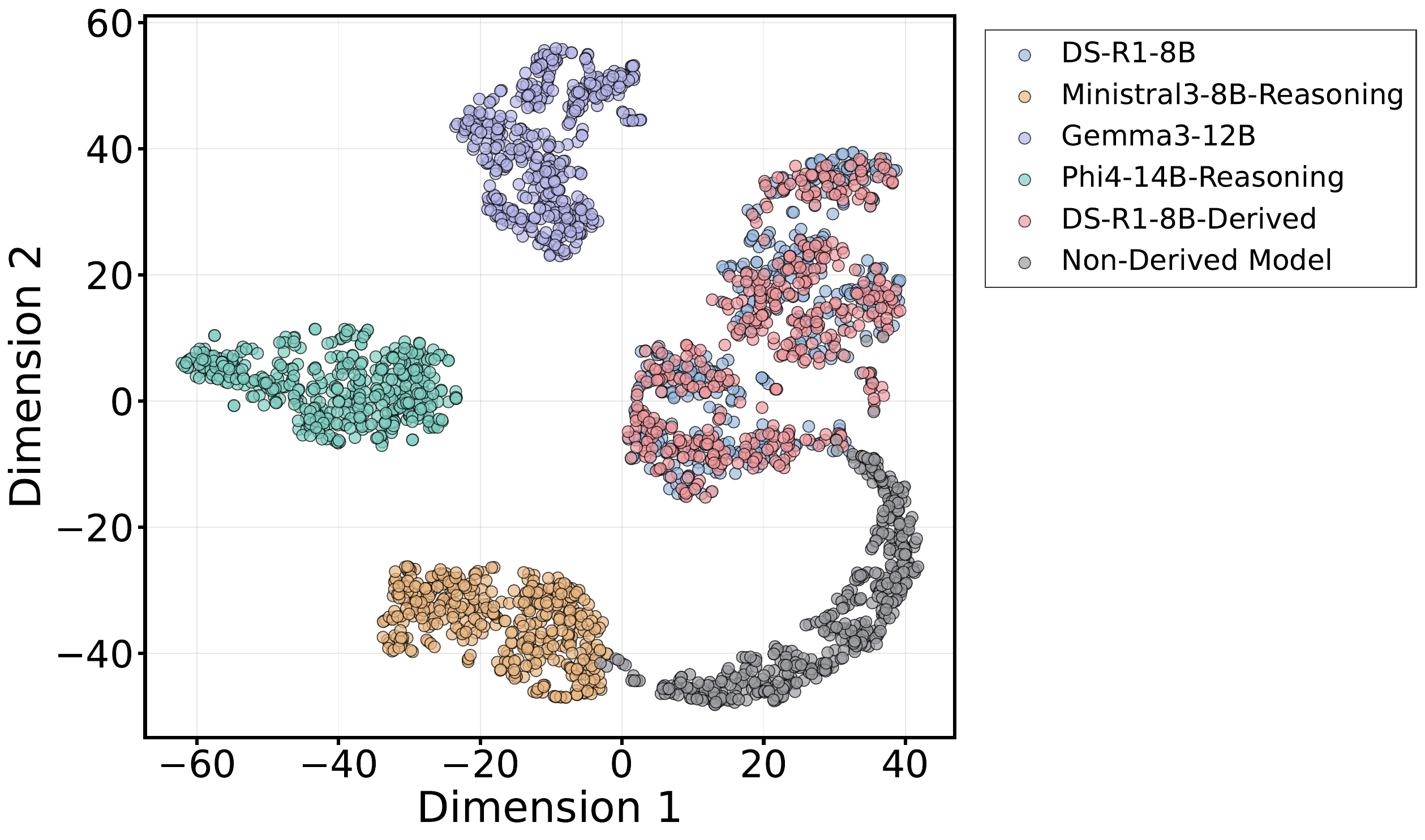} 
\caption{t-SNE plot of embeddings from the trained encoder.}
\label{fig:tsne}
\end{wrapfigure}


\paragraph{Analysis of Probability Traces.}
To explain these performance differences, we visualize token-level probabilities in Figure~\ref{fig:prob_all}.
Our analysis focuses on two key criteria: (1) whether the source model is distinguishable from the reference models, and (2) whether the derived model remains aligned with the source.
In the RAG and quantization settings, both criteria are well satisfied. The probability traces clearly separate the source from the references, while the derived models remain close to the source.
Although individual token probabilities fluctuate, their overall patterns preserve model-specific characteristics.
This separation enables \texttt{TOPF} to identify model provenance without relying on exact textual matches.
In the fine-tuning setting, the source remains distinguishable from the references, but the derived model shows a slight shift away from the source.
This reduced similarity is likely caused by parameter updates during adaptation, which change the model's probability distribution.
The effect is more pronounced in model merging, where the separation between the source and reference models is weakened.
Since merging combines parameters from different models, source-specific probability patterns are diluted, making ownership verification more challenging.


\paragraph{Embedding Visualization.}
Figure~\ref{fig:tsne} visualizes the encoder embeddings using t-SNE~\citep{vandermaaten2008tsne}.
Source and reference models form distinct clusters, indicating that the encoder learns model-specific representations.
Derived models cluster near their corresponding source models, whereas non-derived models tend to lie between source and reference clusters.
The compactness of these clusters further suggests that the learned representations suppress prompt- and sampling-induced variations while preserving provenance-related information.
This structure aligns with our assumption that a suspicious model is considered derived only when it is relatively closer to the source model $M_s$ than to any reference model.
Together, these results confirm that \texttt{TOPF} captures ownership information in the learned embedding space.

\subsection{Robustness Analysis}




\begin{wraptable}[10]{r}{0.6\textwidth}
\centering
\setlength{\tabcolsep}{1pt}
\caption{Impact of the number of reference models on \texttt{TOPF}.}
\label{tab:num_models}
\resizebox{\linewidth}{!}{%
\begin{tabular}{@{}C{0.7cm}
    C{1.4cm}C{1.4cm}
    C{1.4cm}C{1.4cm}
    C{1.4cm}C{1.4cm}@{}}
\toprule
\multirow{2}{*}{Num.} & \multicolumn{2}{c}{DS-R1-8B} 
 & \multicolumn{2}{c}{Qwen3-8B} 
 & \multicolumn{2}{c}{GPT-OSS-20B} \\

\cmidrule(lr){2-3} 
\cmidrule(lr){4-5} 
\cmidrule(lr){6-7}

& AUC & TPR@3\% 
& AUC & TPR@3\% 
& AUC & TPR@3\% \\ 
\midrule

1 & 0.980 & 0.944 & 0.996 & 0.981 & 0.937 & 0.922 \\
2 & 0.946 & 0.944 & 0.939 & 0.815 & 0.988 & 0.941 \\
3 & 0.946 & 0.926 & 0.961 & 0.889 & 0.987 & 0.902 \\

\bottomrule
\end{tabular}%
}
\end{wraptable}

\paragraph{Number of Reference Models.}
Table~\ref{tab:num_models} reports the impact of the number of reference models. Overall, the results vary slightly with different reference set sizes, and using three reference models yields stable performance across settings. On DS-R1-8B, AUC remains 0.946 when increasing the number of references from two to three, while TPR@3\% changes from 0.944 to 0.926. These results indicate that \texttt{TOPF} is not sensitive to the number of reference models. A possible explanation is that the contrastive formulation relies on relative similarity, so a small and diverse reference set is sufficient to form a stable decision boundary.




\begin{wraptable}[12]{r}{0.6\textwidth}
\centering
\setlength{\tabcolsep}{1pt}
\caption{Impact of probe domains on \texttt{TOPF}.}
\label{tab:probe_type}
\resizebox{\linewidth}{!}{%
\begin{tabular}{@{}L{1.2cm}
    C{1.4cm}C{1.4cm}
    C{1.4cm}C{1.4cm}
    C{1.4cm}C{1.4cm}@{}}
\toprule
\multirow{2}{*}{Type} & \multicolumn{2}{c}{DS-R1-8B} 
 & \multicolumn{2}{c}{Qwen3-8B} 
 & \multicolumn{2}{c}{GPT-OSS-20B} \\

\cmidrule(lr){2-3} 
\cmidrule(lr){4-5} 
\cmidrule(lr){6-7}

& AUC & TPR@3\% 
& AUC & TPR@3\% 
& AUC & TPR@3\% \\ 
\midrule

GSM8K    & 0.950 & 0.944 & 0.962 & 0.907 & 0.992 & 0.980 \\
HotpotQA & 0.946 & 0.926 & 0.937 & 0.796 & 0.978 & 0.902 \\
LogiQA   & 0.950 & 0.944 & 0.881 & 0.778 & 0.975 & 0.824 \\
MBPP     & 0.948 & 0.889 & 0.933 & 0.852 & 0.985 & 0.941 \\
XSum     & 0.942 & 0.833 & 0.985 & 0.907 & 0.930 & 0.745 \\
All      & 0.946 & 0.926 & 0.961 & 0.889 & 0.987 & 0.902 \\

\bottomrule
\end{tabular}%
}
\end{wraptable}

\paragraph{Probe Domains.}
Table~\ref{tab:probe_type} examines the effect of probe domains on \texttt{TOPF} using five representative tasks: GSM8K~\citep{cobbe2021training} (math solving), HotpotQA~\citep{yang2018hotpotqa} (multi-hop QA), LogiQA~\citep{liu2020logiqa} (logical reasoning), MBPP~\citep{austin2021program} (code generation), and XSum~\citep{narayan2018dont} (abstractive summarization).
GSM8K yields the strongest results, especially in TPR@3\%.
Overall, these results show that \texttt{TOPF} is robust to the probe domain.
Unlike GCG-based methods, \texttt{TOPF} uses natural prompts, avoiding optimized queries that are easily detected by perplexity-based filters.

\begin{wrapfigure}[14]{r}{0.60\textwidth}
    \centering
    \begin{subfigure}{0.485\linewidth}
        \centering
        \includegraphics[width=\linewidth]{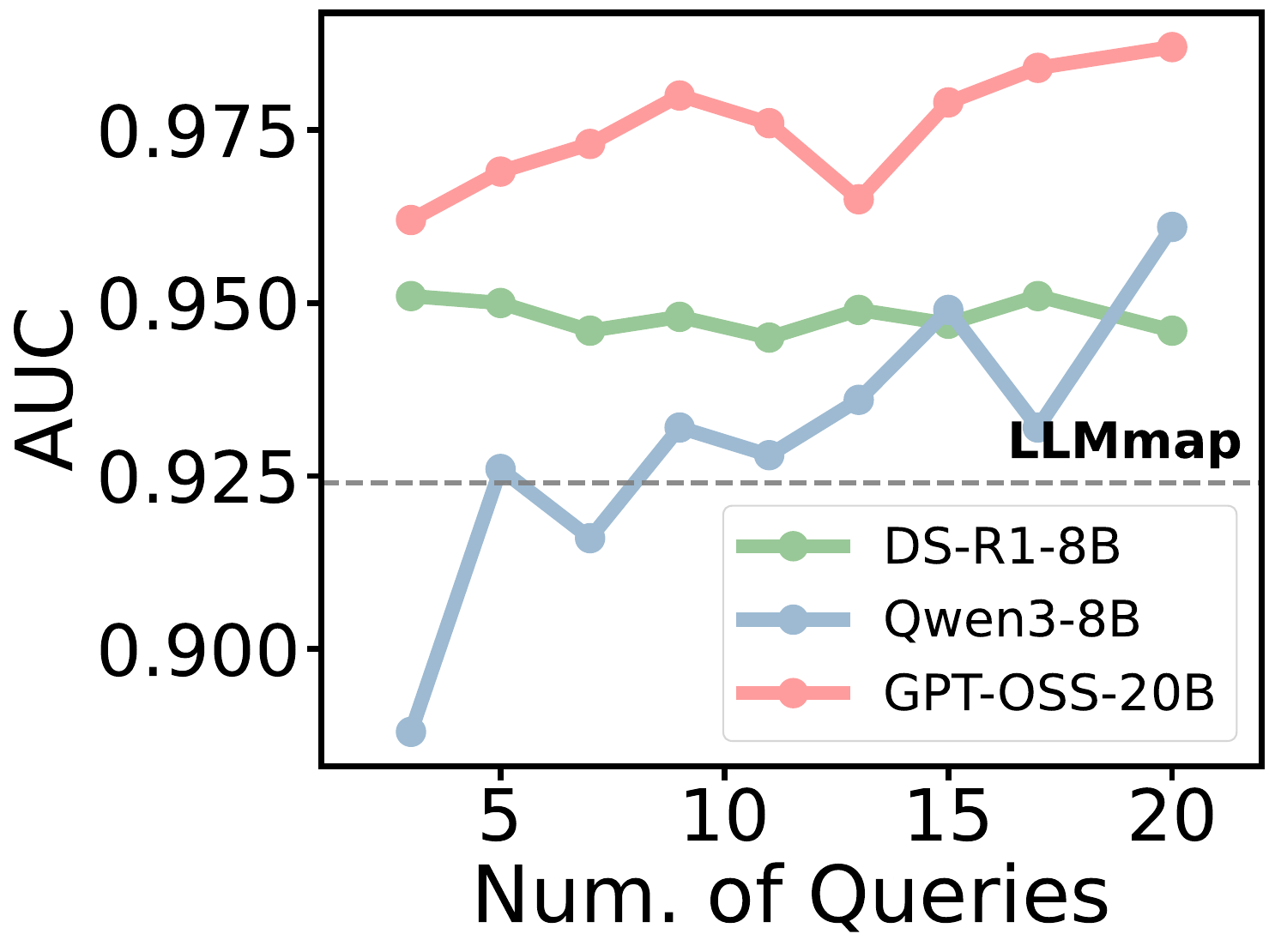}
    \end{subfigure}
    \hspace{0.03\linewidth}
    \begin{subfigure}{0.465\linewidth}
        \centering
        \includegraphics[width=\linewidth]{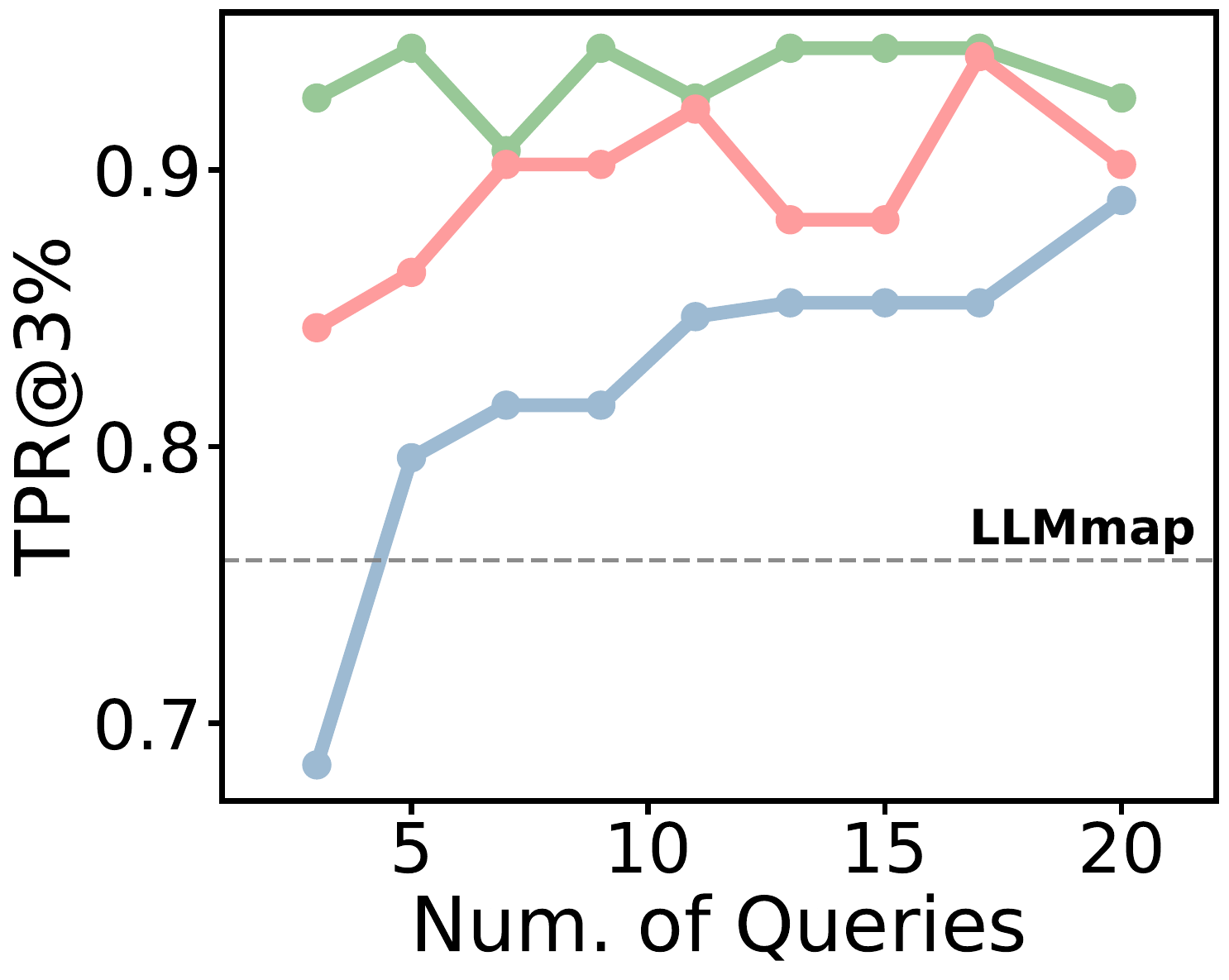}
    \end{subfigure}
    \caption{Impact of the number of queries on \texttt{TOPF}. The left shows AUC and the right shows TPR@3\%. The dashed line indicates the performance of LLMmap on DS-R1-8B.}
    \label{fig:num_queries}
\end{wrapfigure}

\paragraph{Number of Queries.}
Figure~\ref{fig:num_queries} shows the effect of the number of queries issued to the suspicious model. Across all source models, AUC and TPR@3\% improve as the number of queries increases, indicating that more queries provide a more stable estimate. For example, on Qwen3-8B, AUC rises from 0.889 with 3 queries to 0.961 with 20 queries, while TPR@3\% increases from 0.685 to 0.889. Notably, \texttt{TOPF} consistently outperforms the LLMmap baseline even with few queries. These results indicate that \texttt{TOPF} benefits from more queries while remaining effective even under constrained query budgets, which is critical for practical black-box ownership verification.